\documentclass[10pt,twocolumn,letterpaper]{article}

\usepackage{cvpr}              %

\usepackage{graphicx}
\usepackage{amsmath}
\usepackage{amssymb}
\usepackage{booktabs}
\usepackage{pifont}%
\usepackage{xcolor}
\usepackage{mathabx}
\usepackage{enumitem}

\newcommand{\imagic}[0]{\textit{Imagic}}
\newcommand{\editbench}[0]{\textit{TEdBench}}

\newcommand{\vx}[0]{\mathbf{x}}
\newcommand{\vy}[0]{\mathbf{y}}

\newcommand{\ve}[0]{\mathbf{e}}
\newcommand{\veps}[0]{\boldsymbol{\epsilon}}
\newcommand{\mI}[0]{\mathbf{I}}
\newcommand{\bbR}[0]{\mathbb{R}}

\newcommand{\bbE}[0]{\mathbb{E}}
\newcommand{\gN}[0]{\mathcal{N}}
\newcommand{\loss}[0]{\mathcal{L}}

\newcommand{\brs}[1]{\left[ #1 \right]}

\newcommand{\norm}[1]{\left\Vert #1 \right\Vert}

\usepackage[pagebackref,breaklinks,colorlinks]{hyperref}

\usepackage[capitalize]{cleveref}
\crefname{section}{Sec.}{Secs.}
\Crefname{section}{Section}{Sections}
\Crefname{table}{Table}{Tables}
\crefname{table}{Tab.}{Tabs.}

\def\cvprPaperID{1568} %
\def\confName{CVPR}
\def\confYear{2023}

\begin{document}

\title{%
Imagic: Text-Based Real Image Editing with Diffusion Models
}

\author{
\begin{tabular}{cccc}
    Bahjat Kawar$^{*\ 1, 2}$ &
    Shiran Zada$^{*\ 1}$ &
    Oran Lang$^{1}$ &
    Omer Tov$^{1}$ \\
    Huiwen Chang$^{1}$ &
    Tali Dekel$^{1, 3}$ &
    Inbar Mosseri$^{1}$ &
    Michal Irani$^{1, 3}$
\end{tabular} \\
\small{$^1$Google Research \qquad \qquad $^2$Technion \qquad \qquad $^3$Weizmann Institute of Science}
}

\twocolumn[{%
\renewcommand\twocolumn[2][]{#1}%
\maketitle%
\vspace{-0.65cm}	
\centering \centering
\includegraphics[width=\textwidth]{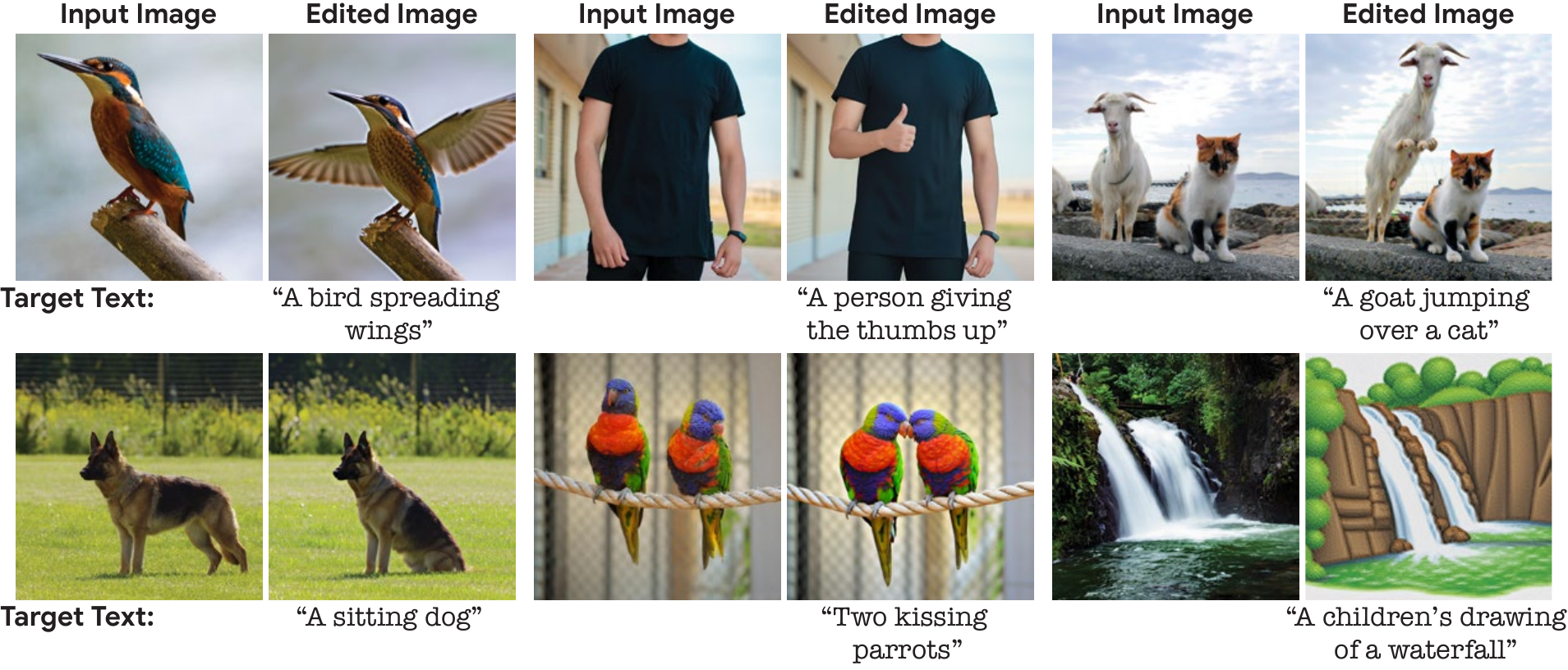}
\captionsetup[figure]{aboveskip=0.0cm}
\captionof{figure}
{\textbf{\imagic\  -- Editing a \underline{single} real image.} {\it Our method can perform various text-based semantic edits on a single real input image, including highly complex non-rigid changes such as posture changes and editing multiple objects.
Here, we show pairs of $1024$$\times$$1024$ input (real) images, and edited outputs with their respective target texts.}
}
\vspace*{0.4cm}
\label{fig:teaser}
}] 

\newcommand\blfootnote[1]{%
  \begingroup
  \renewcommand\thefootnote{}\footnote{#1}%
  \addtocounter{footnote}{-1}%
  \endgroup
}

\blfootnote{$^*$ Equal contribution.}
\blfootnote{The first author performed this work as an intern at Google Research.}
\blfootnote{Project page: \url{https://imagic-editing.github.io/}.}

\begin{abstract}

\vspace*{-0.3cm}
Text-conditioned image editing has recently attracted considerable interest. However, most methods are currently limited to one of the following: specific editing types (\textit{e.g.}, object overlay, style transfer), synthetically generated images, or requiring multiple input images of a common object.
In this paper we demonstrate, for the very first time, the ability to apply complex (e.g., non-rigid) 
text-based %
semantic edits to a single real image. For example, we can change the posture and composition of one or multiple objects inside an image, while preserving its original characteristics. Our method can make a standing dog sit down, %
cause a bird to spread its wings, \textit{etc.} -- each within its 
single high-resolution user-provided natural image.
Contrary to previous work, our proposed method requires only a single input image and a target text (the desired edit). It operates on real images, and does not require any additional inputs (such as image masks or additional views of the object).
Our method, called \imagic, leverages a pre-trained text-to-image diffusion model for this task. It  produces a text embedding that aligns with both the input image and the target text, while fine-tuning the diffusion model to capture the image-specific appearance.
We demonstrate the quality and versatility of 
\imagic\ 
on numerous inputs from various domains, showcasing a plethora of high quality complex semantic 
\mbox{image edits, all within a single unified framework.}
{To better assess performance, we introduce \editbench, a highly challenging image editing benchmark. We conduct a user study, whose findings show that human raters prefer \imagic\  to previous leading editing methods on \editbench.}
\end{abstract}

\vspace*{-0.4cm}
\section{Introduction}
\vspace*{-0.05cm}

\begin{figure*}
    \centering
    \includegraphics[width=0.96\textwidth]{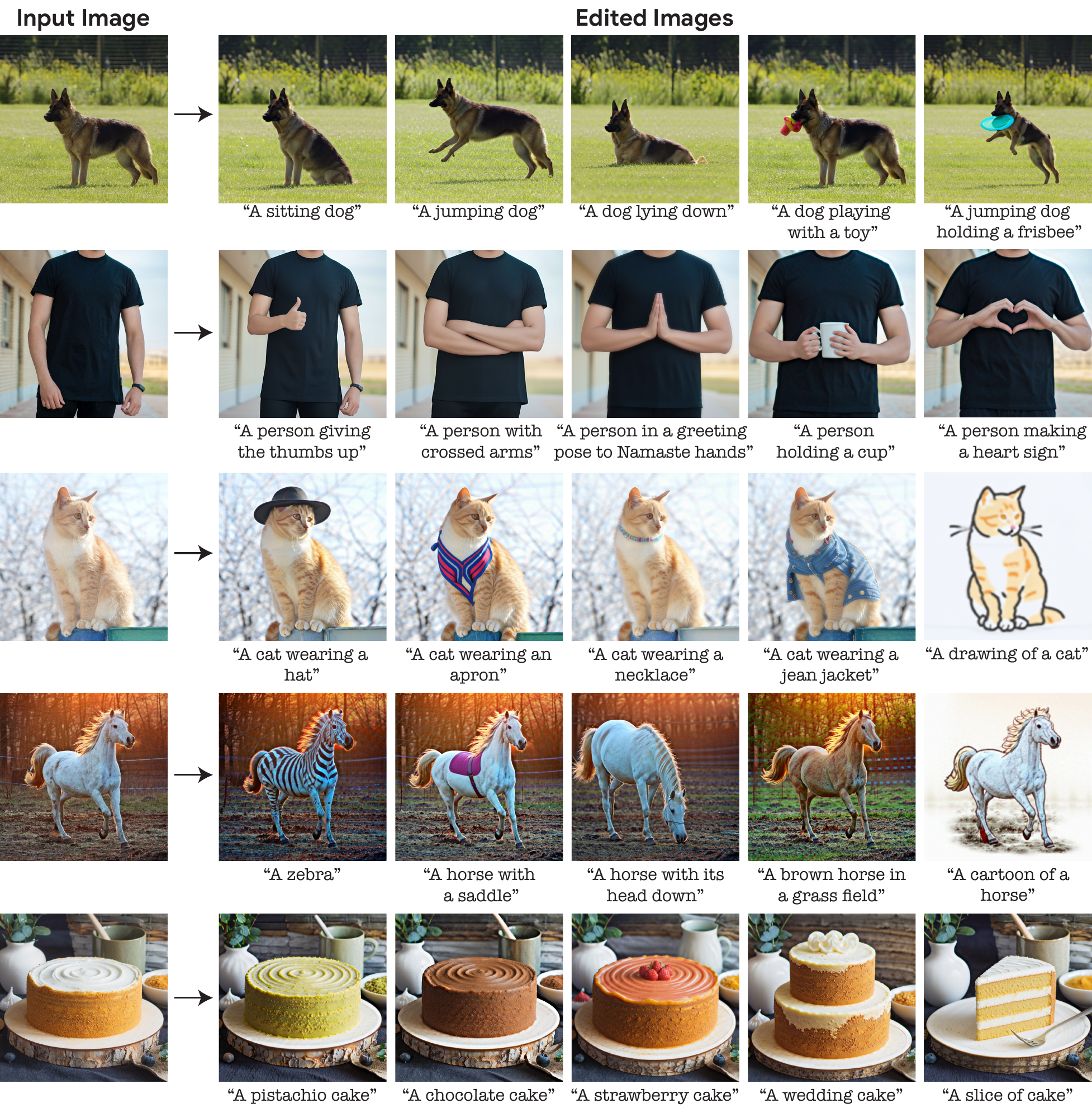}
   \vspace*{-0.1cm}
    \caption{\textbf{Different target texts applied to the same image.} {\it \imagic\  edits the same image differently depending on the input text.}} %
    \label{fig:prompts}
   \vspace*{-0.4cm}
\end{figure*}

Applying non-trivial semantic edits to real photos has long been an interesting task in image processing~\cite{oh2001image}.
It has attracted considerable interest in recent years, enabled by the considerable advancements of deep learning-based systems.
Image editing becomes especially impressive when the desired edit is described by a simple natural language text prompt, since this aligns well with human communication.
Many methods were developed for text-based image editing, showing promising results and continually improving~\cite{bermano2022state, text2live, kim2022diffusionclip}.
However, the current leading methods suffer from, to varying degrees, several drawbacks:
(i) they are limited to a specific set of edits such as painting over the image, adding an object, or transferring style~\cite{kim2022diffusionclip, avrahami2022blended};
(ii) they can operate only on images from a specific domain or synthetically generated images~\cite{patashnik2021styleclip, hertz2022prompt};
or (iii) they require auxiliary inputs in addition to the input image, such as image masks indicating the desired edit location, multiple images of the same subject, or a text describing the original image~\cite{nichol2021glide, dalle2, avrahami2022blended, ruiz2022dreambooth, gal2022textual}.

\begin{figure*}
    \centering
    \includegraphics[width=0.86\textwidth]{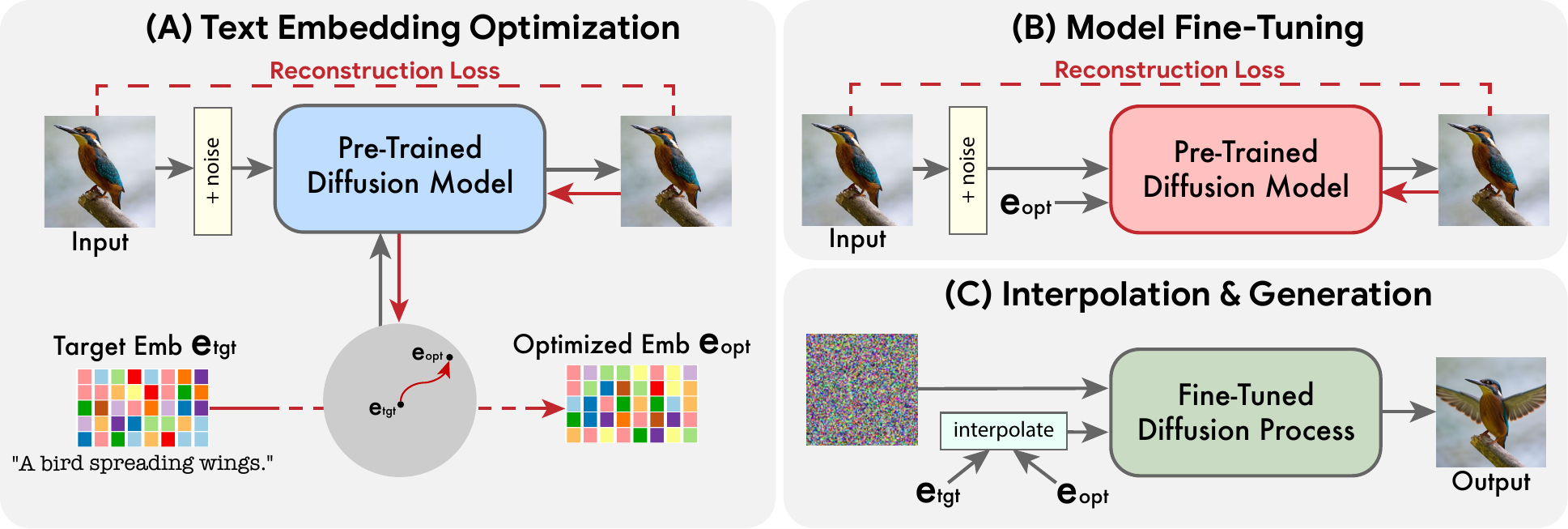}
   \vspace*{-0.25cm}
    \caption{\textbf{Schematic description of \imagic.} {\it Given a real image and a target text prompt: (A) We encode the target text and get the initial text embedding $\ve_{tgt}$, then optimize it to reconstruct the input image, obtaining $\ve_{opt}$; (B) We then fine-tune the generative model to improve fidelity to the input image while fixing $\ve_{opt}$; (C) Finally, we interpolate $\ve_{opt}$ with $\ve_{tgt}$ to generate the final editing result.}
    }
    \label{fig:method}
    \vspace*{-0.45cm}
\end{figure*}

In this paper, we propose a semantic image editing method that mitigates all the above problems.
Given only an input image to be edited and a single text prompt describing the target edit, our method can perform sophisticated non-rigid edits on real high-resolution images.
The resulting image outputs align well with the target text, while preserving the overall background, structure, and composition of the original image.
For example, we can make two parrots kiss or make a person give the thumbs up, as demonstrated in \autoref{fig:teaser}.
Our method, which we call \imagic, provides the first demonstration of text-based semantic editing that applies such sophisticated manipulations to a single real high-resolution image, including editing multiple objects.
In addition, %
\imagic\  can also perform a wide variety of edits, including style changes, color changes, and object additions.

To achieve this feat, we take advantage of the recent success of text-to-image diffusion models~\cite{imagen, dalle2, latent_diffusion}.
Diffusion models are powerful state-of-the-art generative models, capable of high quality image synthesis~\cite{ho2020denoising, dhariwal2021diffusion}. When conditioned on natural language text prompts, they are able to generate images that align well with the requested text.
We adapt them in our work to edit real images instead of synthesizing new ones. We do so in a simple 3-step process, as depicted in \autoref{fig:method}:
We first optimize a text embedding so that it results in images similar to the input image. %
Then, we fine-tune the pre-trained generative diffusion model (conditioned on the optimized embedding) to better reconstruct the input image.
Finally, we linearly interpolate between the target text embedding and the optimized one, resulting in a representation that combines both the input image and the target text. This representation is then passed to the generative diffusion process with the fine-tuned model, which outputs our final edited image.

We conduct several experiments and apply our method on numerous images from various domains.
Our method %
outputs high quality images that both resemble the input image to a high degree, and align well with the target text.
These results showcase the generality, versatility, and quality of \imagic.
We additionally conduct an ablation study, highlighting the effect of each element of our method.
When compared to recent approaches suggested in the literature, \imagic\  exhibits significantly better editing quality and faithfulness to the original image, especially when tasked with sophisticated non-rigid edits.
{This is further supported by a human perceptual evaluation study, where raters strongly prefer \imagic\  over other methods on a novel benchmark called \editbench\ -- Textual Editing Benchmark.
}

\noindent We summarize our main contributions as follows:
\begin{enumerate}[topsep=0pt,itemsep=-1ex,partopsep=1ex,parsep=1ex,leftmargin=*]
\item We present \imagic, the first text-based semantic image editing technique that allows for complex non-rigid edits on a single real input image, while preserving its overall structure and composition.
\item We demonstrate a semantically meaningful linear interpolation between two text embedding sequences, uncovering strong compositional capabilities of text-to-image diffusion models.
\item {We introduce \editbench\  -- a novel and challenging complex image editing benchmark, which enables comparisons of different text-based image editing methods.
    }
\end{enumerate}

\vspace*{-0.1cm}
\section{Related Work}

\begin{figure*}
    \centering
    \includegraphics[width=0.87\textwidth]{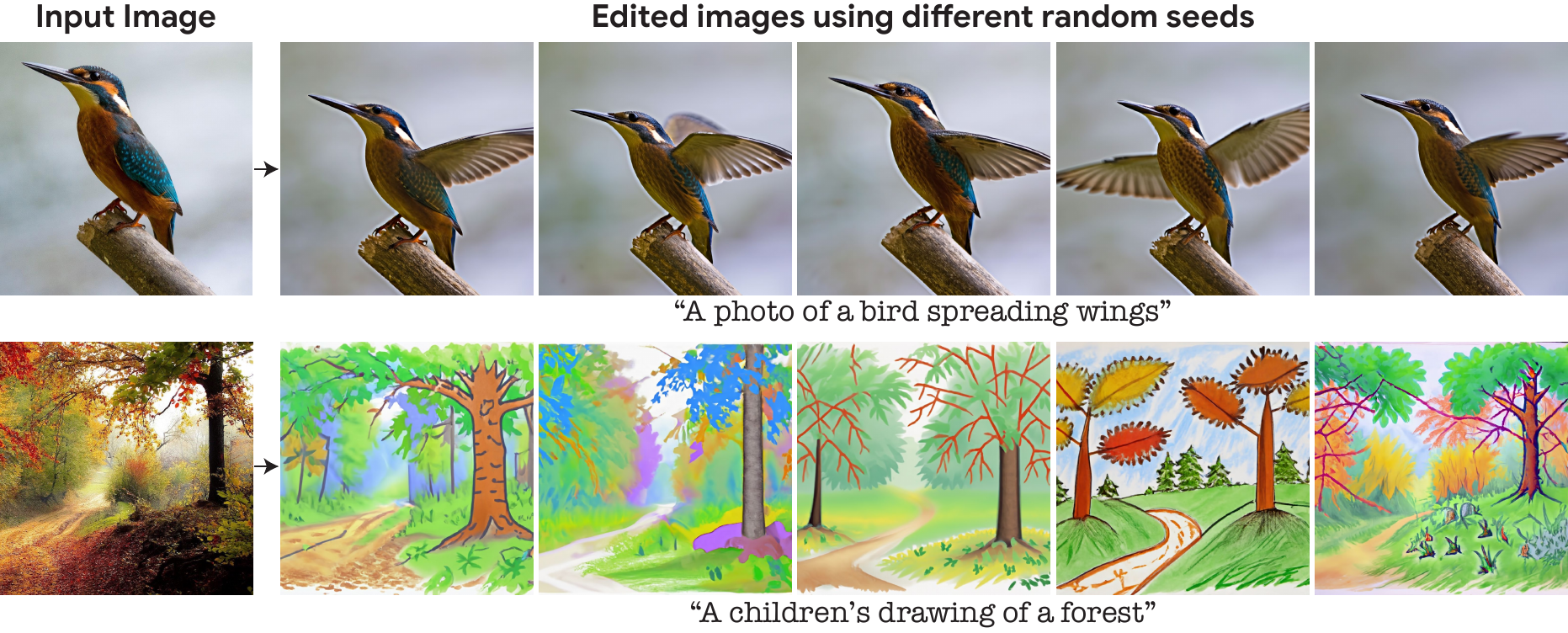}
    \vspace*{-0.3cm}
    \caption{\textbf{Multiple edit options.} {\it \imagic\  utilizes a probabilistic model, enabling it to generate multiple options with different random seeds.}} %
    \label{fig:seeds}
    \vspace*{-0.5cm}
\end{figure*}

Following recent advancements in image synthesis quality \cite{karras2019style,karras2020analyzing,Karras2020ada,karras2021alias}, many works utilized the latent space of pre-trained generative adversarial networks (GANs) to perform a variety of image manipulations \cite{shen2020interpreting, harkonen2020ganspace, lang2021explaining, shen2020closedform,abdal2020styleflow,patashnik2021styleclip}.
{Multiple techniques for applying such manipulations on real images were suggested, including optimization-based methods~\cite{abdal2019image2stylegan,abdal2020image2stylegan++,jahanian2019steerability}, encoder-based methods \cite{richardson2020encoding, tov2021designing,alaluf2021restyle}, and methods adjusting the model per input~\cite{roich2022pivotal, bau2020semantic, alaluf2021hyperstyle, choi2020stargan}.}
In addition to GAN-based methods, some techniques utilize other deep learning-based systems for image editing~\cite{chang2022maskgit, text2live}.

More recently, diffusion models were utilized for similar image manipulation tasks, showcasing remarkable results.
SDEdit~\cite{sdedit} adds intermediate noise to an image (possibly augmented by user-provided brush strokes), then denoises it using a diffusion process conditioned on the desired edit, which is limited to global edits.
{DDIB~\cite{ddib} encodes an input image using DDIM inversion with a source class (or text), and decodes it back conditioned on the target class (or text) to obtain an edited version.}
DiffusionCLIP~\cite{kim2022diffusionclip} utilizes language-vision model gradients, DDIM inversion~\cite{song2020denoising}, and model fine-tuning %
to edit images using a domain-specific diffusion model.
It was also suggested to edit images by synthesizing data in user-provided masks, while keeping the rest of the image intact~\cite{nichol2021glide, avrahami2022blended, choi2021ilvr}.
Liu et al.~\cite{liu2021more} guide a diffusion process with a text and an image, synthesising images similar to the given one, and aligned with the given text.
Hertz et al.~\cite{hertz2022prompt} alter a text-to-image diffusion process by manipulating cross-attention layers, providing more fine-grained control over generated images, and can edit real images in cases where DDIM inversion provides meaningful attention maps.
Textual Inversion~\cite{gal2022textual} and DreamBooth~\cite{ruiz2022dreambooth} synthesize novel views of a given subject given $3$--$5$ images of the subject and a target text (rather than edit a single image), with DreamBooth requiring additional generated images for fine-tuning the models.
In this work, we provide the first text-based semantic image editing tool that operates on a single real image, maintains high fidelity to it, and applies non-rigid edits given a single free-form natural language text prompt.

\section{Imagic: Diffusion-Based Real Image Editing}
\subsection{Preliminaries}
Diffusion models~\cite{vincent2011connection, sohl2015deep, song2019generative, ho2020denoising} are a family of generative models that has recently gained traction, as they advanced the state-of-the-art in image generation~\cite{song2020score, vahdat2021score, dhariwal2021diffusion, kawar2022enhancing}, and have been deployed in various downstream applications such as image restoration~\cite{kawar2022denoising, saharia2022palette}, adversarial purification~\cite{blau2022threat, nie2022diffusion}, image compression~\cite{theis2022lossy}, image classification~\cite{zimmermann2021score}, and others~\cite{wolleb2021diffusion, popov2021grad, kawar2022jpeg, gao2022back, sasaki2021unit, chen2022re}.

The core premise of these models is to initialize with a randomly sampled noise image $\vx_T \sim \gN(0, \mI)$, then iteratively refine it in a controlled fashion, until it is synthesized into a photorealistic image $\vx_0$.
Each intermediate sample $\vx_t$ (for ${t \in \{0, \dots, T\}}$) satisfies
\vspace*{-0.2cm}
\begin{equation}
    \vx_t = \sqrt{\alpha_t} \vx_0 + \sqrt{1 - \alpha_t} \veps_t,
    \label{eq:xt_def}
\vspace*{-0.2cm}
\end{equation}
with ${0 = \alpha_T < \alpha_{T-1} < \dots < \alpha_1 < \alpha_0 = 1}$ being hyperparameters of the diffusion schedule, and $\veps_t \sim \gN(0, \mI)$.
Each refinement step consists of an application of a neural network $f_\theta(\vx_t, t)$ on the current sample $\vx_t$, followed by a random Gaussian noise perturbation, obtaining $\vx_{t-1}$.
The network is trained for a simple denoising objective, aiming for $f_\theta(\vx_t, t) \approx \veps_t$~\cite{sohl2015deep, ho2020denoising}. This leads to a learned image distribution with high fidelity to the target distribution, enabling stellar generative performance.

This method can be generalized for learning conditional distributions -- by conditioning the denoising network on an auxiliary input $\vy$, the network $f_\theta(\vx_t, t, \vy)$ and its resulting diffusion process can faithfully sample from a data distribution conditioned on $\vy$.
The conditioning input $\vy$ can be a low-resolution version of the desired image~\cite{saharia2022image} or a class label~\cite{ho2022cascaded}.
Furthermore, $\vy$ can also be on a text sequence describing the desired image~\cite{imagen, dalle2, latent_diffusion, balaji2022ediffi}.
By incorporating knowledge from large language models (LLMs)~\cite{t5_transformer} or hybrid vision-language models~\cite{clip}, these ~\emph{text-to-image diffusion models} have unlocked a new capability -- users can generate realistic high-resolution images using only a text prompt describing the desired scene.
In all these methods, a low-resolution image is first synthesized using a generative diffusion process, and then it is transformed into a high-resolution one using additional auxiliary models. 
\begin{figure*}
\vspace*{-0.5cm}
    \centering
    \includegraphics[page=1,width=0.86\textwidth]{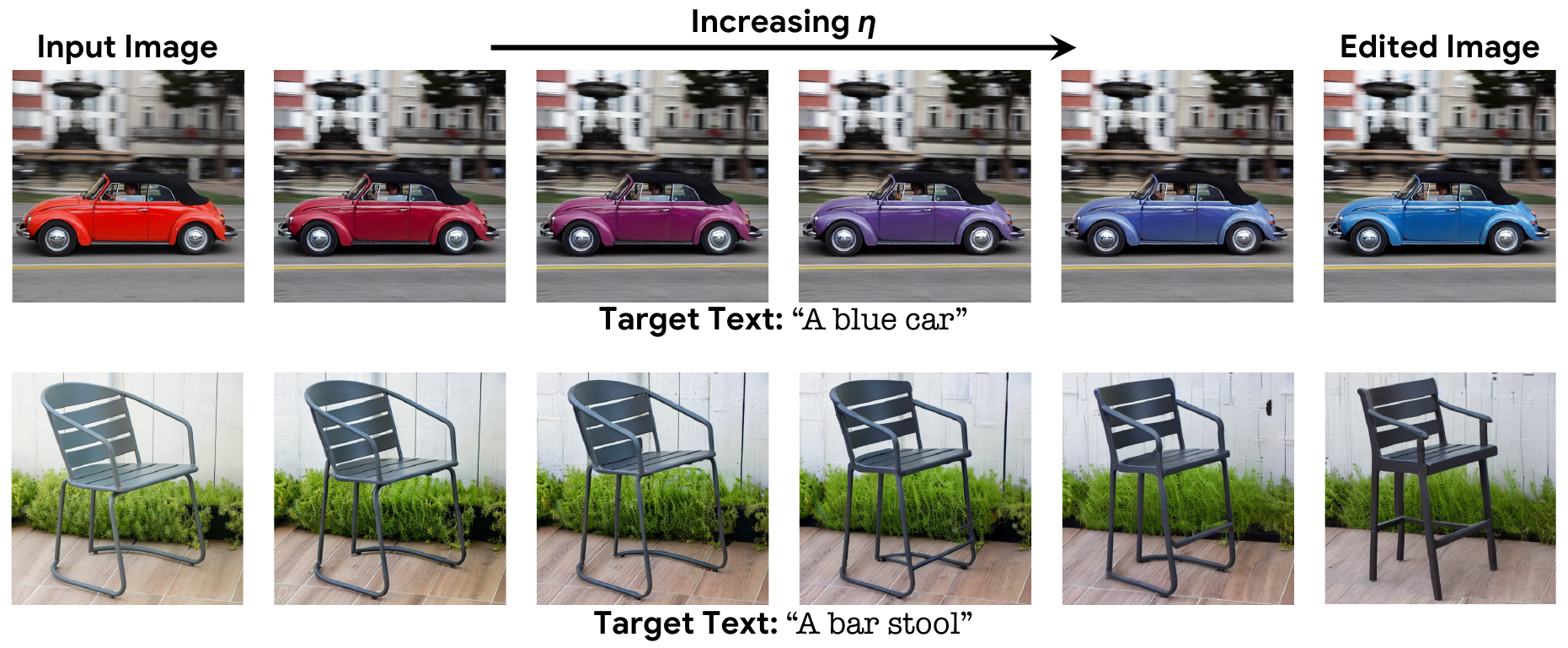}
    \vspace{-0.3cm}
    \caption{{\textbf{Smooth interpolation.} 
    { \it We can smoothly interpolate between the optimized text embedding and the target text embedding, resulting in a gradual editing of the input image toward the required text as $\eta$ increases {(See animated GIFs in supplementary material)}.
    }}}
    \label{fig:stable_difusion_smooth_Interpolation}
   \vspace*{-0.5cm}
\end{figure*}

\subsection{Our Method}

Given an input image $\vx$ and a target text which describes the desired edit, our goal is to edit the image in a way that satisfies the given text, while preserving a maximal amount of detail from $\vx$ (\textit{e.g.}, small details in the background and the identity of the object within the image).
To achieve this feat, we utilize the text embedding layer of the diffusion model to perform semantic manipulations.
Similar to GAN-based approaches~\cite{tov2021designing, roich2022pivotal, patashnik2021styleclip}, we begin by finding meaningful representation which, when fed through the generative process, yields images similar to the input image. We then fine-tune the generative model to better reconstruct the input image and finally manipulate the latent representation to obtain the edit result.

More formally, as depicted in~\autoref{fig:method}, our method  consists of $3$ stages:
(i) we optimize the text embedding to find one that best matches the given image in the vicinity of the target text embedding; 
(ii) we fine-tune the diffusion models to better match the given image;
and (iii) we linearly interpolate between the optimized embedding and the target text embedding, in order to find a point that achieves both fidelity to the input image and target text alignment.
We now turn to describe each step in more detail.

\vspace*{-0.3cm}
\paragraph{Text embedding optimization}
The target text is first passed through a text encoder~\cite{t5_transformer}, which outputs its corresponding text embedding $\ve_{tgt} \in \bbR^{T \times d}$, where $T$ is the number of tokens in the given target text, and $d$ is the token embedding dimension.
We then freeze the parameters of the generative diffusion model $f_{\theta}$, and optimize the target text embedding $\ve_{tgt}$ using the denoising diffusion objective~\cite{ho2020denoising}: %
\vspace*{-0.3cm}
\begin{equation}
    \loss(\vx, \ve, \theta) = 
    \bbE_{t, \veps } \brs{ \norm{\veps - f_{\theta}(\vx_t, t, \ve) }_2^2},
    \label{eq:loss}
\end{equation}
where %
$t$$\sim$${Uniform}[1, T]$, $\vx_t$ is a noisy version of $\vx$ (the input image) obtained using $\veps$$\sim$$\gN(0, \mI)$ and \autoref{eq:xt_def}, and $\theta$ are the pre-trained diffusion model weights.
This results in a text embedding that matches our input image as closely as possible.
We run this process for relatively few steps, in order to remain close to the initial target text embedding, obtaining $\ve_{opt}$.
This proximity enables meaningful linear interpolation in the embedding space, which does not exhibit linear behavior
for  distant embeddings.

\vspace*{-0.3cm}
\paragraph{Model fine-tuning}
Note that the obtained optimized embedding $\ve_{opt}$ does not necessarily lead to the input image $\vx$ exactly when passed through the generative diffusion process, as our optimization runs for a small number of steps (see top left image in  \autoref{fig:finetune_or_not}).
Therefore, in the second stage of our method, we close this gap by optimizing the model parameters $\theta$ using the same loss function presented in \autoref{eq:loss}, while freezing the optimized embedding.
This process shifts the model to fit the input image $\vx$ at the point $\ve_{opt}$. %
In parallel, we fine-tune any auxiliary diffusion models present in the underlying generative method, such as super-resolution models. %
We fine-tune them with the same reconstruction loss, but conditioned on $\ve_{tgt}$, {as they will operate on an edited image.} %
The optimization of these auxiliary models ensures the preservation of high-frequency details from $\vx$ that are not present in the base resolution.
{Empirically, we found that at inference time, inputting  $\mathbf{e}_{tgt}$ to the auxiliary models performs better than using $\mathbf{e}_{opt}$.}

\vspace*{-0.3cm}
\paragraph{Text embedding interpolation}
Since the generative diffusion model was trained to fully recreate the input image $\vx$ at the optimized embedding $\ve_{opt}$, we use it to apply the desired edit by
advancing in the direction of the target text embedding $\ve_{tgt}$. %
More formally, our third stage is a simple linear interpolation between $\ve_{tgt}$ and $\ve_{opt}$.
For a given hyperparameter $\eta \in [0, 1]$, we obtain
\vspace*{-0.11cm}
\begin{equation}
    \bar{\ve} = \eta \cdot \ve_{tgt} + \left(1 - \eta\right) \cdot \ve_{opt},
    \label{eq:interp}
\vspace*{-0.11cm}
\end{equation}
which is the embedding that represents the desired edited image.
We then apply the base generative diffusion process using the fine-tuned model, conditioned on $\bar{\ve}$.
This results in a low-resolution edited image, which is then super-resolved using the fine-tuned auxiliary models, conditioned on the target text.
This generative process outputs our final high-resolution edited image $\bar{\vx}$.

\vspace*{-0.1cm}
\begin{figure}
    \centering
    \hspace*{-3pt}
    \includegraphics[width=0.475\textwidth]{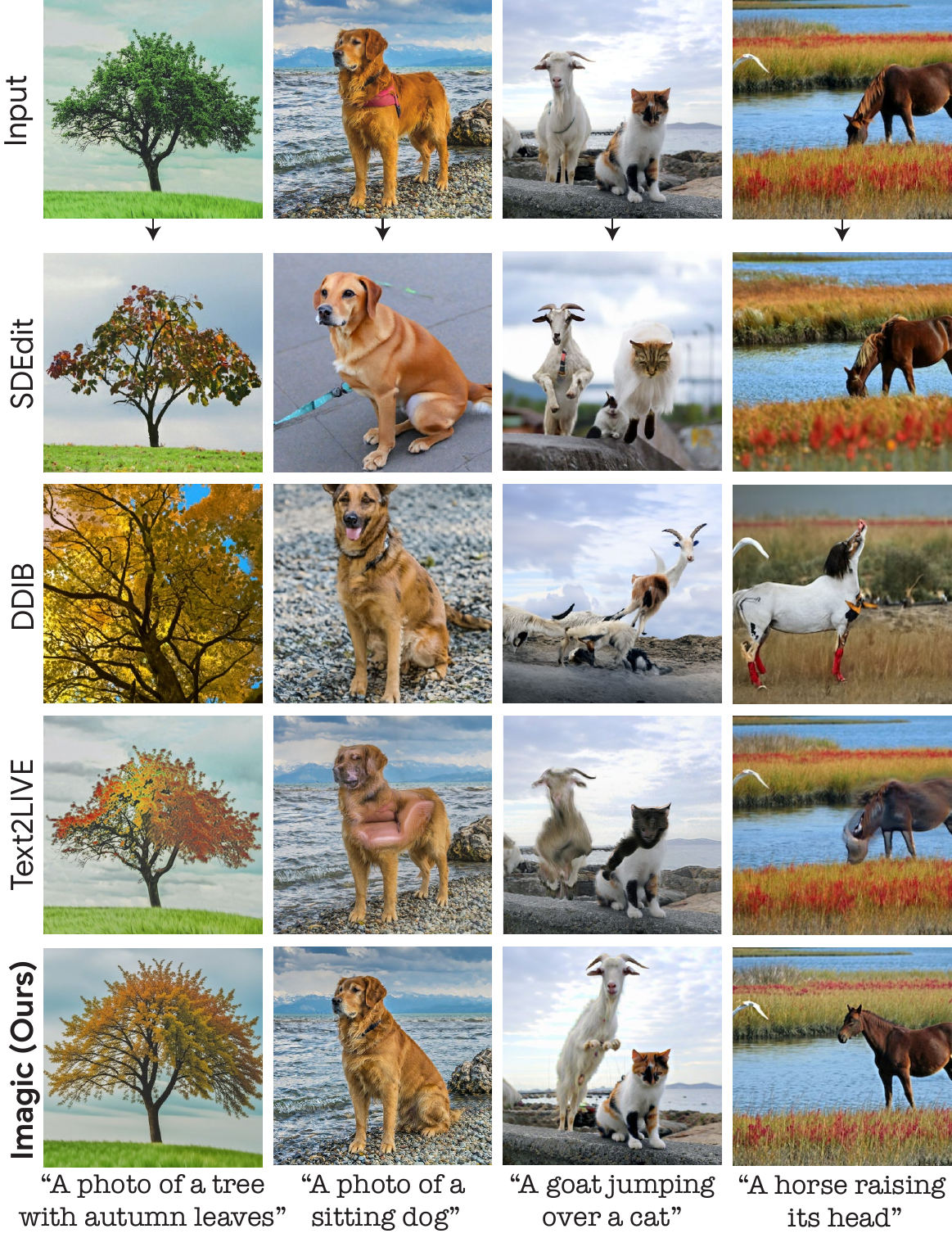}
\vspace*{-0.68cm}
    \caption{\textbf{Method comparison.} \textit{We compare SDEdit~\cite{sdedit}, DDIB~\cite{ddib}, and Text2LIVE~\cite{text2live} to our method. \imagic\  successfully applies the desired edit, %
    while preserving the original image details well.}
    } %
    \label{fig:comparison}
\vspace*{-0.7cm}
\end{figure}

\subsection{Implementation Details}

\vspace*{-0.1cm}
{Our framework is general and can be combined with different generative models. We demonstrate it using two different state-of-the-art text-to-image generative diffusion models: {Imagen}~\cite{imagen} and {Stable Diffusion}~\cite{latent_diffusion}.}

{Imagen}~\cite{imagen} consists of 3 separate text-conditioned~diffusion models:
(i)~a generative diffusion model for $64$$\times$$64$-pixel images; 
(ii)~a super-resolution (SR) diffusion model turning $64$$\times$$64$-pixel images into $256$$\times$$256$ ones; 
and (iii)~another SR model transforming $256$$\times$$256$-pixel images into the $1024$$\times$$1024$ resolution.
By cascading these 3 models~\cite{ho2022cascaded} and using classifier-free guidance~\cite{ho2021classifier}, Imagen constitutes a powerful text-guided image generation scheme.

We optimize the text embedding using the $64$$\times$$64$ diffusion model and the Adam~\cite{adam} optimizer for $100$ steps and a fixed learning rate of $1\text{e}{-3}$.
We then fine-tune the $64$$\times$$64$ diffusion model by continuing Imagen's training for $1500$ steps for our input image, conditioned on the optimized embedding.
In parallel, we also fine-tune the $64$$\times$$64 \rightarrow 256$$\times$$256$ SR diffusion model using the target text embedding and the original image for $1500$ steps, in order to capture high-frequency details from the original image.
We find that fine-tuning the $256$$\times$$256 \rightarrow 1024$$\times$$1024$ model adds little to no effect to the results, therefore we opt to use its pre-trained version conditioned on the target text.
This entire optimization process takes around $8$ minutes per image on two TPUv4 chips.

Afterwards, we interpolate the text embeddings according to \autoref{eq:interp}. Because of the fine-tuning process, using $\eta$$=$$0$ will generate the original image, and as $\eta$ increases, the image will start to align with the target text. 
To maintain both image fidelity and target text alignment, we choose an intermediate $\eta$, usually residing between $0.6$ and $0.8$ (see \autoref{fig:lpips_clip}).
We then generate with Imagen~\cite{imagen} with its provided hyperparameters. We find that using the DDIM~\cite{song2020denoising} sampling scheme generally provides slightly improved results over the more stochastic DDPM scheme.

{
In addition to {Imagen}, we also implement our method with the publicly available {Stable Diffusion} model (based on Latent Diffusion Models~\cite{latent_diffusion}).
This model applies the diffusion process in the latent space (of size $4$$\times$$64$$\times$$64$) of a pre-trained autoencoder, working with $512$$\times$$512$-pixel images.
We apply our method in the latent space as well. We optimize the text embedding for $1000$ steps with a learning rate of $2\text{e}{-3}$ using Adam~\cite{adam}. Then, we fine-tune the diffusion model for $1500$ steps with a learning rate of $5\text{e}{-7}$.
This process takes $7$ minutes on a single Tesla A100 GPU.
}

\vspace*{-0.1cm}
\section{Experiments}
\vspace*{-0.1cm}

\begin{figure*}
    \centering
    \includegraphics[width=0.88\textwidth]{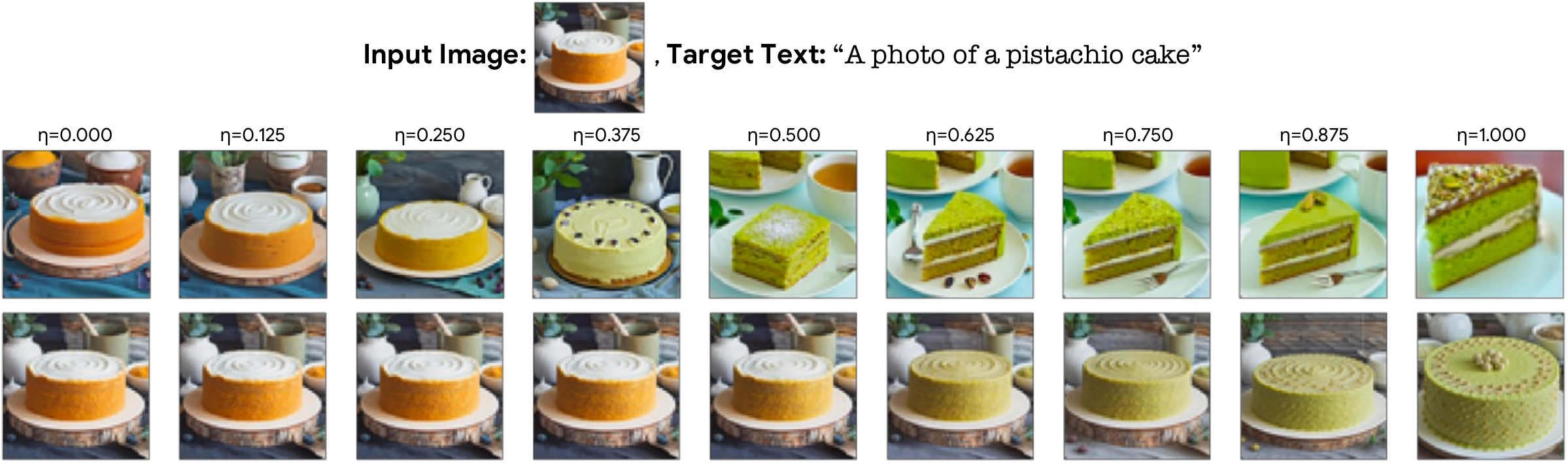}
   \vspace*{-0.35cm}
    \caption{\textbf{Embedding interpolation.} {\it Varying $\eta$ with the same seed, using the pre-trained (top) and fine-tuned (bottom) models.}}
    \label{fig:finetune_or_not}
   \vspace*{-0.25cm}
\end{figure*}

\begin{figure}
     \centering
     \begin{subfigure}[b]{0.155\textwidth}
     \hspace*{-1em}
         \centering
         \includegraphics[width=1.1\textwidth]{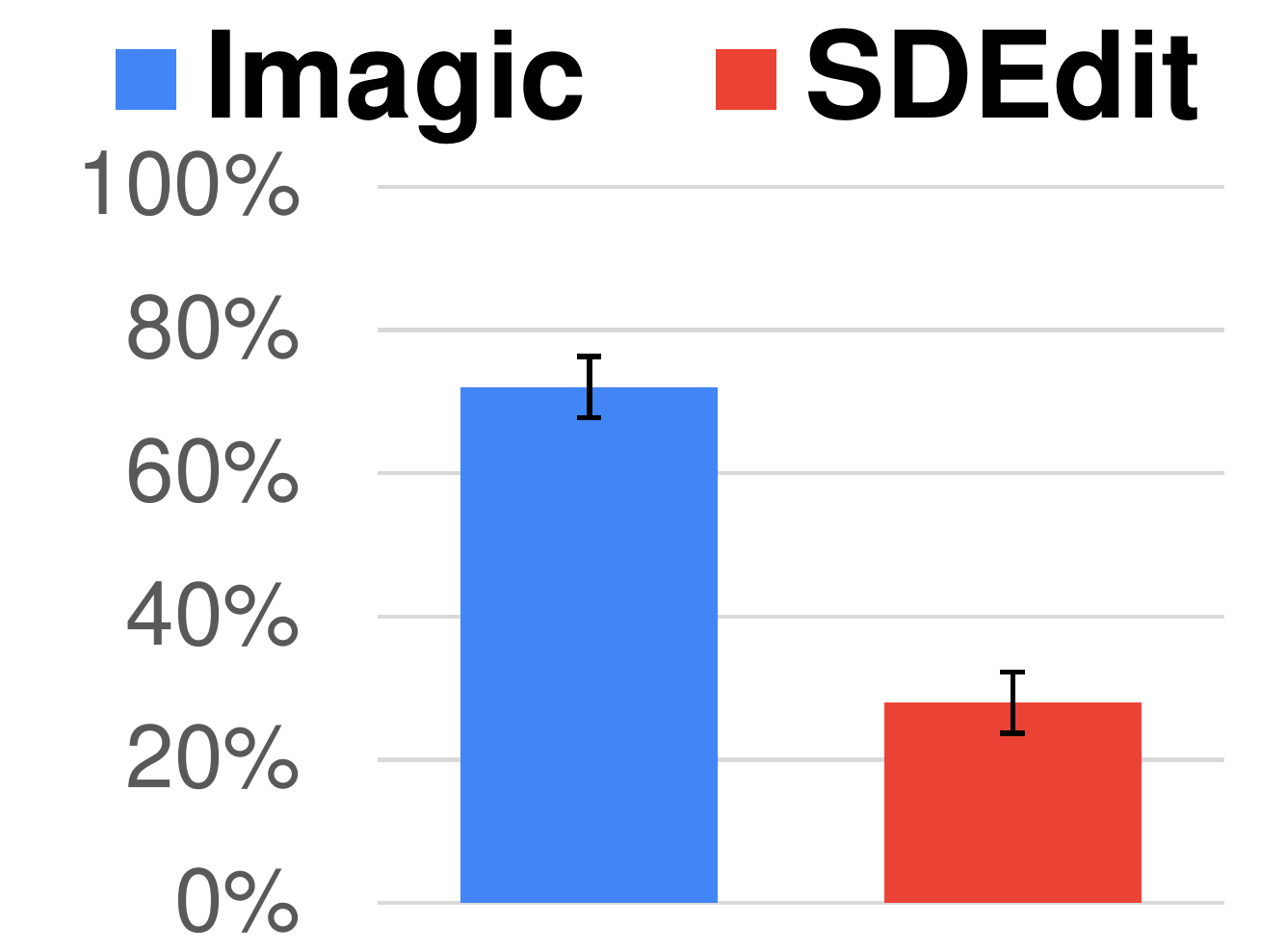}
     \end{subfigure}
     \hfill
     \begin{subfigure}[b]{0.155\textwidth}
     \hspace*{-1em}
         \centering
         \includegraphics[width=1.1\textwidth]{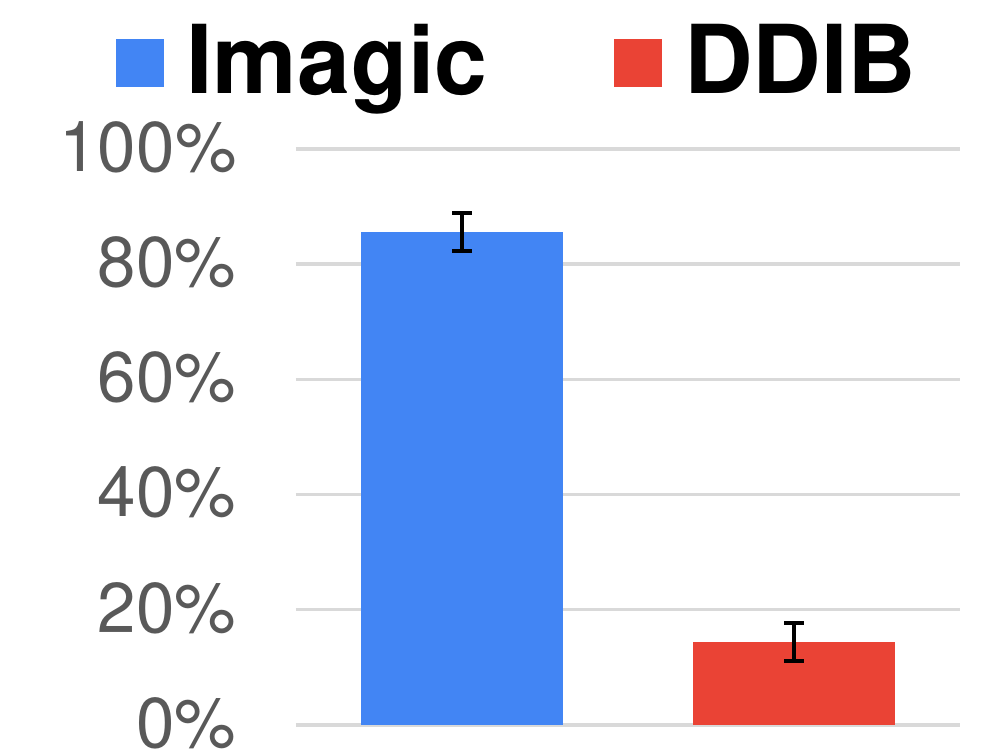}
     \end{subfigure}
     \hfill
     \begin{subfigure}[b]{0.155\textwidth}
     \hspace*{-1em}
         \centering
         \includegraphics[width=1.1\textwidth]{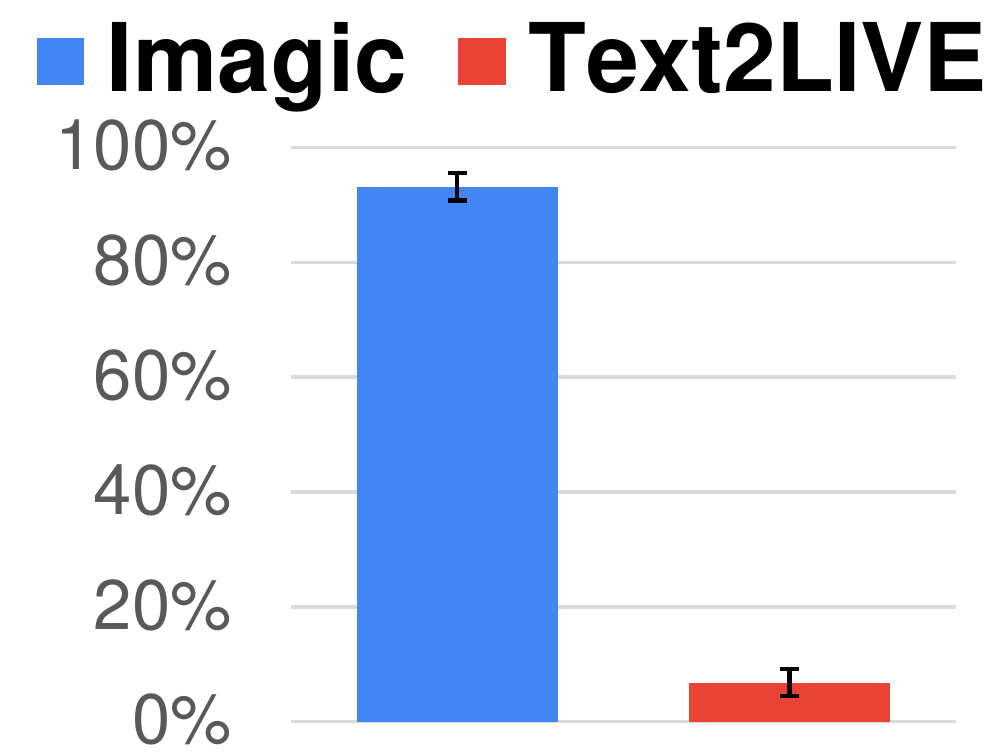}
     \end{subfigure}    
        \vspace*{-0.7cm}
     \caption{\textbf{User study results.} {\textit{Preference rates (with $95\%$ confidence intervals) for image editing quality of \imagic\  over SDEdit~\cite{sdedit}, DDIB~\cite{ddib}, and Text2LIVE~\cite{text2live}.}}}
        \label{fig:user_study}
   \vspace*{-0.5cm}
\end{figure}

\subsection{Qualitative Evaluation}
We applied our method on a multitude of real images from various domains, with simple text prompts describing different editing categories such as: style, appearance, color, posture, and composition.
We collect high-resolution free-to-use images from Unsplash and Pixabay.
After optimization, we generate each edit with $8$ random seeds and choose the best result.
\imagic\  %
is able to apply various editing categories on general input images and texts, as we show in \autoref{fig:teaser} and the supplementary material. %
We experiment with different text prompts for the same image in \autoref{fig:prompts}, showing the versatility of \imagic.
Since the underlying generative diffusion model that we utilize is probabilistic, our method can generate different results for a single image-text pair. We show multiple options for the same edit using different random seeds in \autoref{fig:seeds}, slightly tweaking $\eta$ for each seed.
This stochasticity allows the user to choose among these different options, as natural language text prompts can generally be ambiguous and imprecise.

While we use Imagen~\cite{imagen} in most of our experiments, Imagic is agnostic to the generative model choice.
Thus, we also implement Imagic with Stable Diffusion~\cite{latent_diffusion}.
In \autoref{fig:stable_difusion_smooth_Interpolation} (and in the supplementary material) we show that \imagic\  successfully
performs 
complex non-rigid edits also using Stable Diffusion while 
preserving the image-specific appearance.
Furthermore, \imagic\  (using Stable Diffusion) exhibits smooth semantic interpolation properties as $\eta$ is changed.
We hypothesize that this smoothness property is a byproduct of the diffusion process taking place in a semantic latent space, rather than in the image pixel space.

\subsection{Comparisons}
We compare \imagic\  to the current leading general-purpose techniques that operate on a single input real-world image, and edit it based on a text prompt. %
Namely, we compare our method to Text2LIVE~\cite{text2live}, DDIB~\cite{ddib}, and SDEdit~\cite{sdedit}.  We use Text2LIVE's default provided hyperparameters. We feed it with a text description of the target object (\textit{e.g.}, ``dog'') and one of the desired edit (\textit{e.g.}, ``sitting dog'').
For SDEdit and DDIB, we apply their proposed technique with the same Imagen~\cite{imagen} model and target text prompt that we use. We keep the diffusion hyperparameters from Imagen, and choose the intermediate diffusion timestep for SDEdit independently for each image to achieve the best target text alignment without drastically changing the image contents. For DDIB, we provide an additional source text.

\autoref{fig:comparison} shows editing results of different methods. For SDEdit and \imagic, we sample $8$ images using different random seeds and display the result with the best alignment to both the target text and the input image.
As can be observed, %
our method maintains high fidelity to the input image while aptly performing the desired edits.
When tasked with a complex non-rigid edit such as making a dog sit, our method significantly outperforms previous techniques.
\imagic\  constitutes the first demonstration of such sophisticated text-based edits applied on a single real-world image.
{We verify this claim through a user study in \autoref{sec:user_study}.}

\vspace*{-0.1cm}
\subsection{TEdBench and User Study}
\label{sec:user_study}
\vspace*{-0.1cm}
{Text-based image editing methods are a relatively recent development,
and \imagic\  is the first to apply complex non-rigid edits.
As such, no standard benchmark exists for evaluating  non-rigid text-based image editing.
We introduce \editbench\ (Textual Editing Benchmark), a novel collection of $100$ pairs of input images and target texts describing a desired complex non-rigid edit.
We hope that future research will benefit from \editbench\  as a standardized evaluation set for this task.}

{We quantitatively evaluate \imagic's performance via an extensive human perceptual evaluation study on \editbench, performed using Amazon Mechanical Turk.
Participants were shown an input image and a target text,
and were asked to choose the better editing result from one of two options, using the standard practice of Two-Alternative Forced Choice (2AFC)~\cite{kolkin2019style, park2020swapping, text2live}.
The options to choose from were our result and a baseline result from one of: SDEdit~\cite{sdedit}, DDIB~\cite{ddib}, or Text2LIVE~\cite{text2live}.
In total, we collected $9213$ answers, whose results are summarized in~\autoref{fig:user_study}. As can be seen, evaluators exhibit a strong preference towards our method, with a preference rate of more than $70\%$ across all considered baselines.
See supplementary material for more details about the user study and method implementations.
}

\subsection{Ablation Study}

\begin{figure}
\vspace*{-0.5cm}
  \begin{minipage}[c]{0.52\columnwidth}
    \hspace*{-0.2cm}
    \includegraphics[width=1.00\columnwidth]{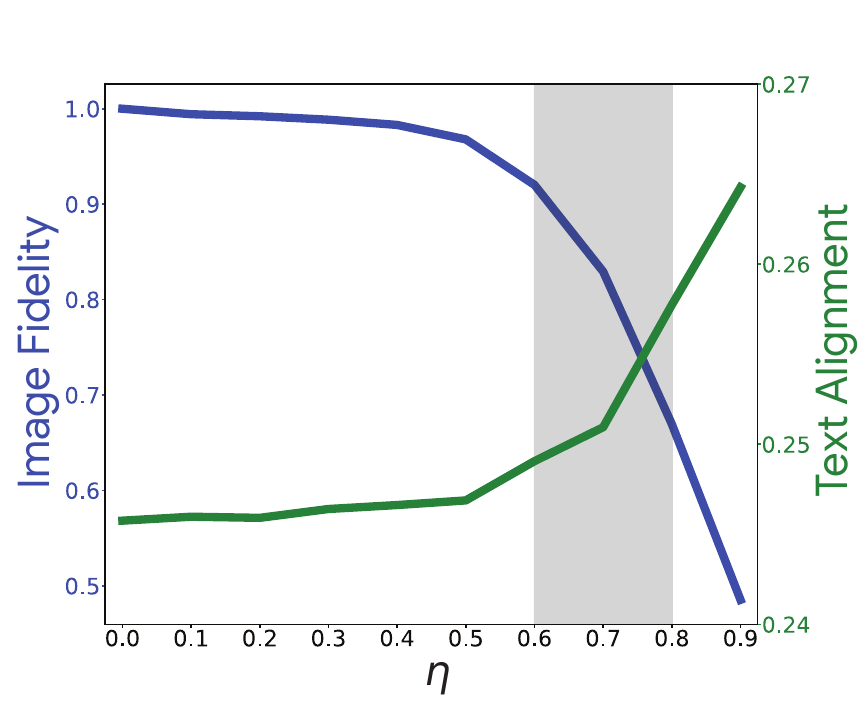}
  \end{minipage}\hfill
  \begin{minipage}[c]{0.48\columnwidth} 
    \hspace*{0.2cm}
    \caption{\textbf{Editability--fidelity tradeoff.} {\it CLIP score (target text alignment) and $1 - $LPIPS (input image fidelity) as functions of $\eta$, averaged over $150$ inputs. Edited images tend to match both the input image and text in the highlighted area. 
    }
\label{fig:lpips_clip}
    } 
  \end{minipage}
   \vspace*{-0.7cm}
\end{figure}

\vspace*{-0.1cm}
\paragraph{Fine-tuning and optimization}
We generate edited images for different $\eta$ values using the pre-trained $64\times64$ diffusion model and our fine-tuned one, in order to gauge the effect of fine-tuning on the output quality. We use the same optimized embedding and random seed, and qualitatively evaluate the results in \autoref{fig:finetune_or_not}.
Without fine-tuning, the scheme does not fully reconstruct the original image at $\eta=0$, and fails to retain the image's details as $\eta$ increases.
In contrast, fine-tuning imposes details from the input image beyond just the optimized embedding, allowing our scheme to retain these details for intermediate values of $\eta$, thereby enabling semantically meaningful linear interpolation.
Thus, we conclude that model fine-tuning is essential for our method's success.
Furthermore, we experiment with the number of text embedding optimization steps
in the supplementary material.
Our findings suggest that optimizing the text embedding with a smaller number of steps limits our editing capabilities, while optimizing for more than $100$ steps yields little to no added value.

\vspace*{-0.3cm}
\paragraph{Interpolation intensity}
As can be observed in \autoref{fig:finetune_or_not}, fine-tuning increases the $\eta$ value at which the model strays from reconstructing the input image.
While the optimal $\eta$ value may vary per input (as different edits require different intensities), we attempt to identify the region in which the edit is best applied.
To that end, we apply our editing scheme with different $\eta$ values, and calculate the outputs' CLIP score~\cite{clip, clipscore} w.r.t. the target text, and their LPIPS score\cite{lpips} w.r.t. the input image subtracted from $1$.
A higher CLIP score indicates better output alignment with the target text, and a higher $1 - $LPIPS indicates higher fidelity to the input image.
We repeat this process for $150$ image-text inputs, and show the average results in \autoref{fig:lpips_clip}.
We observe that for $\eta$ values smaller than $0.4$, outputs are almost identical to the input images.
For $\eta \in [0.6, 0.8]$, the images begin to change (according to LPIPS), and align better with the text (as the CLIP score rises).
Therefore, we identify this area as the most probable for obtaining satisfactory results.
Note that while they provide a good sense of text or image alignment on average, CLIP score and LPIPS are imprecise measures that rely on neural network backbones, and their values noticeably differ for each different input image-text pair.
As such, they are not suited for reliably choosing $\eta$ for each input in an automatic way, nor can they faithfully assess an editing method's performance.

\subsection{Limitations}

\begin{figure}
    \centering
    \includegraphics[width=0.4\textwidth]{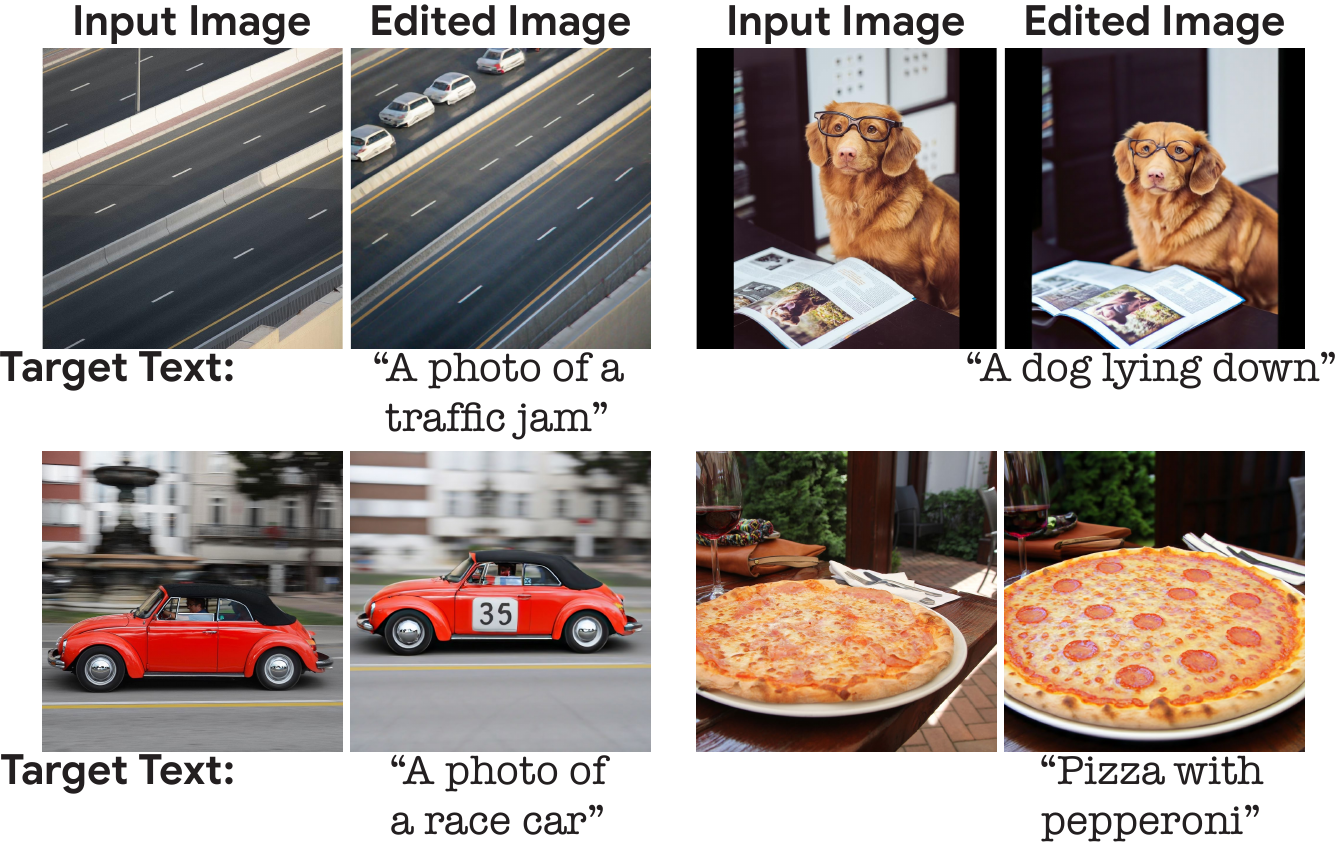}
   \vspace*{-0.3cm}
    \caption{\textbf{Failure cases.} 
{\it Insufficient consistency with  
the target text (top), or changes in camera viewing angle  (bottom).}}
    \label{fig:failures}
   \vspace*{-0.5cm}
\end{figure}

We identify two {main} failure cases of our method: 
In some cases, the desired edit is applied very subtly {(if at all)}, therefore not aligning well with the target text.
In other cases, the edit is applied well, but it affects extrinsic image details such as zoom or camera angle.
We show examples of these two failure cases in the first and second row of \autoref{fig:failures}, respectively.
When the edit is not applied strongly enough, increasing $\eta$ usually achieves the desired result, but it sometimes leads to a significant loss of original image details (for all tested random seeds) in a handful of cases.
As for zoom and camera angle changes, these usually occur before the desired edit takes place, as we progress from a low $\eta$ value to a large one, which makes circumventing them difficult.
We demonstrate this %
in
the supplementary material, and include additional failure cases in \editbench\  as well.

These limitations can possibly be mitigated by optimizing the text embedding or the diffusion model differently, or by incorporating cross-attention control akin to Hertz et al.~\cite{hertz2022prompt}. We leave those options for future work.
Also, since our method relies on a pre-trained text-to-image diffusion model, it inherits the model's generative limitations and biases.
Therefore, unwanted artifacts are produced when the desired edit involves generating failure cases of the underlying model.
For instance, Imagen is known to show substandard generative performance on human faces~\cite{imagen}.
Additionally, the optimization required by Imagic (and other editing methods~\cite{text2live}) is slow, and may hinder their direct deployment in user-facing applications.

\vspace*{-0.1cm}
\section{Conclusions and Future Work}

\vspace*{-0.1cm}
We propose a novel image editing method called \imagic. Our method accepts a single image and a simple text prompt describing the desired edit, and aims to apply this edit while preserving a maximal amount of details from the image.
To that end, we utilize a pre-trained text-to-image diffusion model and use it to find a text embedding that represents the input image. Then, we fine-tune the diffusion model to fit the image better, and finally we linearly interpolate between the embedding representing the image and the target text embedding, obtaining a semantically meaningful mixture of them.
This enables our scheme to provide edited images using the interpolated embedding.
Contrary to other editing methods, our approach can produce sophisticated non-rigid edits that may alter the pose, geometry, and/or composition of objects within the image as requested, in addition to simpler edits such as style or color.
It requires the user to provide only a single image and a simple target text prompt, without the need for additional auxiliary inputs such as image masks.

Our future work may focus on further improving the method's fidelity to the input image and identity preservation, as well as its sensitivity to random seeds and to the interpolation parameter $\eta$. Another intriguing research direction would be the development of an automated method for choosing $\eta$ for each requested edit.

\vspace*{-0.3cm}
\paragraph{Societal Impact} 
Our method aims
to enable complex editing of real world images using textual descriptions of the target edit. 
As such, it is prone to societal biases of the underlying text-based generative models, albeit to a lesser extent than purely generative methods since we rely mostly on the input image for editing. 
However, as with other approaches that use generative models for image editing, such techniques might be used by malicious parties for
synthesizing fake imagery to mislead viewers. To mitigate this, further research on the identification of synthetically edited or generated 
content is needed.

\newpage
{\small
\bibliographystyle{ieee_fullname}
\bibliography{egbib}

\begin{thebibliography}{10}\itemsep=-1pt

\bibitem{abdal2019image2stylegan}
Rameen Abdal, Yipeng Qin, and Peter Wonka.
\newblock Image2stylegan: How to embed images into the stylegan latent space?
\newblock In {\em Proceedings of the IEEE international conference on computer
  vision}, pages 4432--4441, 2019.

\bibitem{abdal2020image2stylegan++}
Rameen Abdal, Yipeng Qin, and Peter Wonka.
\newblock Image2stylegan++: How to edit the embedded images?
\newblock In {\em Proceedings of the IEEE/CVF Conference on Computer Vision and
  Pattern Recognition}, pages 8296--8305, 2020.

\bibitem{abdal2020styleflow}
Rameen Abdal, Peihao Zhu, Niloy Mitra, and Peter Wonka.
\newblock Styleflow: Attribute-conditioned exploration of stylegan-generated
  images using conditional continuous normalizing flows, 2020.

\bibitem{alaluf2021restyle}
Yuval Alaluf, Or Patashnik, and Daniel Cohen-Or.
\newblock Restyle: A residual-based stylegan encoder via iterative refinement.
\newblock In {\em Proceedings of the IEEE/CVF International Conference on
  Computer Vision (ICCV)}, October 2021.

\bibitem{alaluf2021hyperstyle}
Yuval Alaluf, Omer Tov, Ron Mokady, Rinon Gal, and Amit~H Bermano.
\newblock Hyperstyle: Stylegan inversion with hypernetworks for real image
  editing.
\newblock {\em arXiv preprint arXiv:2111.15666}, 2021.

\bibitem{avrahami2022blended}
Omri Avrahami, Dani Lischinski, and Ohad Fried.
\newblock Blended diffusion for text-driven editing of natural images.
\newblock In {\em Proceedings of the IEEE/CVF Conference on Computer Vision and
  Pattern Recognition}, pages 18208--18218, 2022.

\bibitem{balaji2022ediffi}
Yogesh Balaji, Seungjun Nah, Xun Huang, Arash Vahdat, Jiaming Song, Karsten
  Kreis, Miika Aittala, Timo Aila, Samuli Laine, Bryan Catanzaro, et~al.
\newblock {eDiff-I}: Text-to-image diffusion models with an ensemble of expert
  denoisers.
\newblock {\em arXiv preprint arXiv:2211.01324}, 2022.

\bibitem{text2live}
Omer Bar-Tal, Dolev Ofri-Amar, Rafail Fridman, Yoni Kasten, and Tali Dekel.
\newblock {Text2LIVE:} text-driven layered image and video editing.
\newblock {\em arXiv preprint arXiv:2204.02491}, 2022.

\bibitem{bau2020semantic}
David Bau, Hendrik Strobelt, William Peebles, Jonas Wulff, Bolei Zhou, Jun-Yan
  Zhu, and Antonio Torralba.
\newblock Semantic photo manipulation with a generative image prior.
\newblock {\em arXiv preprint arXiv:2005.07727}, 2020.

\bibitem{bermano2022state}
Amit~H Bermano, Rinon Gal, Yuval Alaluf, Ron Mokady, Yotam Nitzan, Omer Tov,
  Oren Patashnik, and Daniel Cohen-Or.
\newblock State-of-the-art in the architecture, methods and applications of
  stylegan.
\newblock In {\em Computer Graphics Forum}, volume~41, pages 591--611. Wiley
  Online Library, 2022.

\bibitem{blau2022threat}
Tsachi Blau, Roy Ganz, Bahjat Kawar, Alex Bronstein, and Michael Elad.
\newblock Threat model-agnostic adversarial defense using diffusion models.
\newblock {\em arXiv preprint arXiv:2207.08089}, 2022.

\bibitem{chang2022maskgit}
Huiwen Chang, Han Zhang, Lu Jiang, Ce Liu, and William~T Freeman.
\newblock Maskgit: Masked generative image transformer.
\newblock In {\em Proceedings of the IEEE/CVF Conference on Computer Vision and
  Pattern Recognition}, pages 11315--11325, 2022.

\bibitem{chen2022re}
Wenhu Chen, Hexiang Hu, Chitwan Saharia, and William~W Cohen.
\newblock Re-imagen: Retrieval-augmented text-to-image generator.
\newblock {\em arXiv preprint arXiv:2209.14491}, 2022.

\bibitem{choi2021ilvr}
Jooyoung Choi, Sungwon Kim, Yonghyun Jeong, Youngjune Gwon, and Sungroh Yoon.
\newblock {ILVR}: Conditioning method for denoising diffusion probabilistic
  models.
\newblock In {\em 2021 IEEE/CVF International Conference on Computer Vision
  (ICCV)}, pages 14347--14356. IEEE, 2021.

\bibitem{choi2020stargan}
Yunjey Choi, Youngjung Uh, Jaejun Yoo, and Jung-Woo Ha.
\newblock {StarGAN} v2: Diverse image synthesis for multiple domains.
\newblock In {\em Proceedings of the IEEE/CVF conference on computer vision and
  pattern recognition}, pages 8188--8197, 2020.

\bibitem{dhariwal2021diffusion}
Prafulla Dhariwal and Alexander Nichol.
\newblock Diffusion models beat gans on image synthesis.
\newblock {\em Advances in Neural Information Processing Systems},
  34:8780--8794, 2021.

\bibitem{gal2022textual}
Rinon Gal, Yuval Alaluf, Yuval Atzmon, Or Patashnik, Amit~H. Bermano, Gal
  Chechik, and Daniel Cohen-Or.
\newblock An image is worth one word: Personalizing text-to-image generation
  using textual inversion, 2022.

\bibitem{gao2022back}
Jin Gao, Jialing Zhang, Xihui Liu, Trevor Darrell, Evan Shelhamer, and Dequan
  Wang.
\newblock Back to the source: Diffusion-driven test-time adaptation.
\newblock {\em arXiv preprint arXiv:2207.03442}, 2022.

\bibitem{harkonen2020ganspace}
Erik H{\"a}rk{\"o}nen, Aaron Hertzmann, Jaakko Lehtinen, and Sylvain Paris.
\newblock Ganspace: Discovering interpretable gan controls.
\newblock {\em arXiv preprint arXiv:2004.02546}, 2020.

\bibitem{hertz2022prompt}
Amir Hertz, Ron Mokady, Jay Tenenbaum, Kfir Aberman, Yael Pritch, and Daniel
  Cohen-Or.
\newblock Prompt-to-prompt image editing with cross attention control, 2022.

\bibitem{clipscore}
Jack Hessel, Ari Holtzman, Maxwell Forbes, Ronan~Le Bras, and Yejin Choi.
\newblock Clipscore: A reference-free evaluation metric for image captioning.
\newblock {\em arXiv preprint arXiv:2104.08718}, 2021.

\bibitem{ho2020denoising}
Jonathan Ho, Ajay Jain, and Pieter Abbeel.
\newblock Denoising diffusion probabilistic models.
\newblock {\em Advances in Neural Information Processing Systems},
  33:6840--6851, 2020.

\bibitem{ho2022cascaded}
Jonathan Ho, Chitwan Saharia, William Chan, David~J Fleet, Mohammad Norouzi,
  and Tim Salimans.
\newblock Cascaded diffusion models for high fidelity image generation.
\newblock {\em Journal of Machine Learning Research}, 23(47):1--33, 2022.

\bibitem{ho2021classifier}
Jonathan Ho and Tim Salimans.
\newblock Classifier-free diffusion guidance.
\newblock In {\em NeurIPS 2021 Workshop on Deep Generative Models and
  Downstream Applications}, 2021.

\bibitem{jahanian2019steerability}
Ali Jahanian, Lucy Chai, and Phillip Isola.
\newblock On the" steerability" of generative adversarial networks.
\newblock In {\em International Conference on Learning Representations}, 2019.

\bibitem{Karras2020ada}
Tero Karras, Miika Aittala, Janne Hellsten, Samuli Laine, Jaakko Lehtinen, and
  Timo Aila.
\newblock Training generative adversarial networks with limited data.
\newblock In {\em Proc. NeurIPS}, 2020.

\bibitem{karras2021alias}
Tero Karras, Miika Aittala, Samuli Laine, Erik H{\"a}rk{\"o}nen, Janne
  Hellsten, Jaakko Lehtinen, and Timo Aila.
\newblock Alias-free generative adversarial networks.
\newblock {\em Advances in Neural Information Processing Systems}, 34, 2021.

\bibitem{karras2019style}
Tero Karras, Samuli Laine, and Timo Aila.
\newblock A style-based generator architecture for generative adversarial
  networks.
\newblock In {\em Proceedings of the IEEE conference on computer vision and
  pattern recognition}, pages 4401--4410, 2019.

\bibitem{karras2020analyzing}
Tero Karras, Samuli Laine, Miika Aittala, Janne Hellsten, Jaakko Lehtinen, and
  Timo Aila.
\newblock Analyzing and improving the image quality of stylegan.
\newblock In {\em Proceedings of the IEEE/CVF Conference on Computer Vision and
  Pattern Recognition}, pages 8110--8119, 2020.

\bibitem{kawar2022denoising}
Bahjat Kawar, Michael Elad, Stefano Ermon, and Jiaming Song.
\newblock Denoising diffusion restoration models.
\newblock In {\em Advances in Neural Information Processing Systems}, 2022.

\bibitem{kawar2022enhancing}
Bahjat Kawar, Roy Ganz, and Michael Elad.
\newblock Enhancing diffusion-based image synthesis with robust classifier
  guidance.
\newblock {\em arXiv preprint arXiv:2208.08664}, 2022.

\bibitem{kawar2022jpeg}
Bahjat Kawar, Jiaming Song, Stefano Ermon, and Michael Elad.
\newblock {JPEG} artifact correction using denoising diffusion restoration
  models.
\newblock {\em arXiv preprint arXiv:2209.11888}, 2022.

\bibitem{kim2022diffusionclip}
Gwanghyun Kim, Taesung Kwon, and Jong~Chul Ye.
\newblock Diffusionclip: Text-guided diffusion models for robust image
  manipulation.
\newblock In {\em Proceedings of the IEEE/CVF Conference on Computer Vision and
  Pattern Recognition}, pages 2426--2435, 2022.

\bibitem{adam}
Diederik~P. Kingma and Jimmy Ba.
\newblock Adam: {A} method for stochastic optimization.
\newblock In Yoshua Bengio and Yann LeCun, editors, {\em 3rd International
  Conference on Learning Representations, {ICLR} 2015, San Diego, CA, USA, May
  7-9, 2015, Conference Track Proceedings}, 2015.

\bibitem{kolkin2019style}
Nicholas Kolkin, Jason Salavon, and Gregory Shakhnarovich.
\newblock Style transfer by relaxed optimal transport and self-similarity.
\newblock In {\em Proceedings of the IEEE/CVF Conference on Computer Vision and
  Pattern Recognition}, pages 10051--10060, 2019.

\bibitem{lang2021explaining}
Oran Lang, Yossi Gandelsman, Michal Yarom, Yoav Wald, Gal Elidan, Avinatan
  Hassidim, William~T. Freeman, Phillip Isola, Amir Globerson, Michal Irani,
  and Inbar Mosseri.
\newblock Explaining in style: Training a gan to explain a classifier in
  stylespace.
\newblock In {\em Proceedings of the IEEE/CVF International Conference on
  Computer Vision}, pages 693--702, 2021.

\bibitem{liu2021more}
Xihui Liu, Dong~Huk Park, Samaneh Azadi, Gong Zhang, Arman Chopikyan, Yuxiao
  Hu, Humphrey Shi, Anna Rohrbach, and Trevor Darrell.
\newblock More control for free! image synthesis with semantic diffusion
  guidance.
\newblock {\em arXiv preprint arXiv:2112.05744}, 2021.

\bibitem{sdedit}
Chenlin Meng, Yutong He, Yang Song, Jiaming Song, Jiajun Wu, Jun-Yan Zhu, and
  Stefano Ermon.
\newblock {SDEdit:} guided image synthesis and editing with stochastic
  differential equations.
\newblock In {\em International Conference on Learning Representations}, 2021.

\bibitem{nichol2021glide}
Alex Nichol, Prafulla Dhariwal, Aditya Ramesh, Pranav Shyam, Pamela Mishkin,
  Bob McGrew, Ilya Sutskever, and Mark Chen.
\newblock Glide: Towards photorealistic image generation and editing with
  text-guided diffusion models.
\newblock {\em arXiv preprint arXiv:2112.10741}, 2021.

\bibitem{nie2022diffusion}
Weili Nie, Brandon Guo, Yujia Huang, Chaowei Xiao, Arash Vahdat, and Anima
  Anandkumar.
\newblock Diffusion models for adversarial purification.
\newblock In {\em International Conference on Machine Learning (ICML)}, 2022.

\bibitem{oh2001image}
Byong~Mok Oh, Max Chen, Julie Dorsey, and Fr{\'e}do Durand.
\newblock Image-based modeling and photo editing.
\newblock In {\em Proceedings of the 28th annual conference on Computer
  graphics and interactive techniques}, pages 433--442, 2001.

\bibitem{park2020swapping}
Taesung Park, Jun-Yan Zhu, Oliver Wang, Jingwan Lu, Eli Shechtman, Alexei
  Efros, and Richard Zhang.
\newblock Swapping autoencoder for deep image manipulation.
\newblock {\em Advances in Neural Information Processing Systems},
  33:7198--7211, 2020.

\bibitem{patashnik2021styleclip}
Or Patashnik, Zongze Wu, Eli Shechtman, Daniel Cohen-Or, and Dani Lischinski.
\newblock Styleclip: Text-driven manipulation of stylegan imagery.
\newblock {\em arXiv preprint arXiv:2103.17249}, 2021.

\bibitem{popov2021grad}
Vadim Popov, Ivan Vovk, Vladimir Gogoryan, Tasnima Sadekova, and Mikhail
  Kudinov.
\newblock Grad-tts: A diffusion probabilistic model for text-to-speech.
\newblock In {\em International Conference on Machine Learning}, pages
  8599--8608. PMLR, 2021.

\bibitem{clip}
Alec Radford, Jong~Wook Kim, Chris Hallacy, Aditya Ramesh, Gabriel Goh,
  Sandhini Agarwal, Girish Sastry, Amanda Askell, Pamela Mishkin, Jack Clark,
  et~al.
\newblock Learning transferable visual models from natural language
  supervision.
\newblock In {\em International Conference on Machine Learning}, pages
  8748--8763. PMLR, 2021.

\bibitem{t5_transformer}
Colin Raffel, Noam Shazeer, Adam Roberts, Katherine Lee, Sharan Narang, Michael
  Matena, Yanqi Zhou, Wei Li, and Peter~J Liu.
\newblock Exploring the limits of transfer learning with a unified text-to-text
  transformer.
\newblock {\em Journal of Machine Learning Research}, 21:1--67, 2020.

\bibitem{dalle2}
Aditya Ramesh, Prafulla Dhariwal, Alex Nichol, Casey Chu, and Mark Chen.
\newblock Hierarchical text-conditional image generation with clip latents.
\newblock {\em arXiv preprint arXiv:2204.06125}, 2022.

\bibitem{richardson2020encoding}
Elad Richardson, Yuval Alaluf, Or Patashnik, Yotam Nitzan, Yaniv Azar, Stav
  Shapiro, and Daniel Cohen-Or.
\newblock Encoding in style: a stylegan encoder for image-to-image translation.
\newblock {\em arXiv preprint arXiv:2008.00951}, 2020.

\bibitem{roich2022pivotal}
Daniel Roich, Ron Mokady, Amit~H Bermano, and Daniel Cohen-Or.
\newblock Pivotal tuning for latent-based editing of real images.
\newblock {\em ACM Transactions on Graphics (TOG)}, 42(1):1--13, 2022.

\bibitem{latent_diffusion}
Robin Rombach, Andreas Blattmann, Dominik Lorenz, Patrick Esser, and Bj{\"o}rn
  Ommer.
\newblock High-resolution image synthesis with latent diffusion models.
\newblock In {\em Proceedings of the IEEE/CVF Conference on Computer Vision and
  Pattern Recognition}, pages 10684--10695, 2022.

\bibitem{ruiz2022dreambooth}
Nataniel Ruiz, Yuanzhen Li, Varun Jampani, Yael Pritch, Michael Rubinstein, and
  Kfir Aberman.
\newblock {DreamBooth:} fine tuning text-to-image diffusion models for
  subject-driven generation.
\newblock {\em arXiv preprint arxiv:2208.12242}, 2022.

\bibitem{saharia2022palette}
Chitwan Saharia, William Chan, Huiwen Chang, Chris Lee, Jonathan Ho, Tim
  Salimans, David Fleet, and Mohammad Norouzi.
\newblock Palette: Image-to-image diffusion models.
\newblock In {\em ACM SIGGRAPH 2022 Conference Proceedings}, pages 1--10, 2022.

\bibitem{imagen}
Chitwan Saharia, William Chan, Saurabh Saxena, Lala Li, Jay Whang, Emily
  Denton, Seyed Kamyar~Seyed Ghasemipour, Burcu~Karagol Ayan, S~Sara Mahdavi,
  Rapha~Gontijo Lopes, et~al.
\newblock Photorealistic text-to-image diffusion models with deep language
  understanding.
\newblock In {\em Advances in Neural Information Processing Systems}, 2022.

\bibitem{saharia2022image}
Chitwan Saharia, Jonathan Ho, William Chan, Tim Salimans, David~J Fleet, and
  Mohammad Norouzi.
\newblock Image super-resolution via iterative refinement.
\newblock {\em IEEE Transactions on Pattern Analysis and Machine Intelligence},
  2022.

\bibitem{sasaki2021unit}
Hiroshi Sasaki, Chris~G Willcocks, and Toby~P Breckon.
\newblock Unit-ddpm: Unpaired image translation with denoising diffusion
  probabilistic models.
\newblock {\em arXiv preprint arXiv:2104.05358}, 2021.

\bibitem{shen2020interpreting}
Yujun Shen, Jinjin Gu, Xiaoou Tang, and Bolei Zhou.
\newblock Interpreting the latent space of gans for semantic face editing.
\newblock In {\em Proceedings of the IEEE/CVF Conference on Computer Vision and
  Pattern Recognition}, pages 9243--9252, 2020.

\bibitem{shen2020closedform}
Yujun Shen and Bolei Zhou.
\newblock Closed-form factorization of latent semantics in gans.
\newblock {\em arXiv preprint arXiv:2007.06600}, 2020.

\bibitem{sohl2015deep}
Jascha Sohl-Dickstein, Eric Weiss, Niru Maheswaranathan, and Surya Ganguli.
\newblock Deep unsupervised learning using nonequilibrium thermodynamics.
\newblock In {\em International Conference on Machine Learning}, pages
  2256--2265. PMLR, 2015.

\bibitem{song2020denoising}
Jiaming Song, Chenlin Meng, and Stefano Ermon.
\newblock Denoising diffusion implicit models.
\newblock In {\em International Conference on Learning Representations}, 2020.

\bibitem{song2019generative}
Yang Song and Stefano Ermon.
\newblock Generative modeling by estimating gradients of the data distribution.
\newblock {\em Advances in Neural Information Processing Systems}, 32, 2019.

\bibitem{song2020score}
Yang Song, Jascha Sohl-Dickstein, Diederik~P Kingma, Abhishek Kumar, Stefano
  Ermon, and Ben Poole.
\newblock Score-based generative modeling through stochastic differential
  equations.
\newblock In {\em International Conference on Learning Representations}, 2020.

\bibitem{ddib}
Xuan Su, Jiaming Song, Chenlin Meng, and Stefano Ermon.
\newblock Dual diffusion implicit bridges for image-to-image translation.
\newblock {\em arXiv preprint arXiv:2203.08382}, 2022.

\bibitem{theis2022lossy}
Lucas Theis, Tim Salimans, Matthew~D Hoffman, and Fabian Mentzer.
\newblock Lossy compression with {G}aussian diffusion.
\newblock {\em arXiv preprint arXiv:2206.08889}, 2022.

\bibitem{tov2021designing}
Omer Tov, Yuval Alaluf, Yotam Nitzan, Or Patashnik, and Daniel Cohen-Or.
\newblock Designing an encoder for stylegan image manipulation.
\newblock {\em arXiv preprint arXiv:2102.02766}, 2021.

\bibitem{vahdat2021score}
Arash Vahdat, Karsten Kreis, and Jan Kautz.
\newblock Score-based generative modeling in latent space.
\newblock {\em Advances in Neural Information Processing Systems},
  34:11287--11302, 2021.

\bibitem{vincent2011connection}
Pascal Vincent.
\newblock A connection between score matching and denoising autoencoders.
\newblock {\em Neural computation}, 23(7):1661--1674, 2011.

\bibitem{wolleb2021diffusion}
Julia Wolleb, Robin Sandk{\"u}hler, Florentin Bieder, Philippe Valmaggia, and
  Philippe~C Cattin.
\newblock Diffusion models for implicit image segmentation ensembles.
\newblock {\em arXiv preprint arXiv:2112.03145}, 2021.

\bibitem{lpips}
Richard Zhang, Phillip Isola, Alexei~A Efros, Eli Shechtman, and Oliver Wang.
\newblock The unreasonable effectiveness of deep features as a perceptual
  metric.
\newblock In {\em Proceedings of the IEEE conference on computer vision and
  pattern recognition}, pages 586--595, 2018.

\bibitem{zimmermann2021score}
Roland~S Zimmermann, Lukas Schott, Yang Song, Benjamin~A Dunn, and David~A
  Klindt.
\newblock Score-based generative classifiers.
\newblock {\em arXiv preprint arXiv:2110.00473}, 2021.

\end{thebibliography}
}

\clearpage
\onecolumn
\appendix

\section{Additional Results}
\vfill
\begin{figure*}[h]
    \centering
    \includegraphics[width=0.97\textwidth]{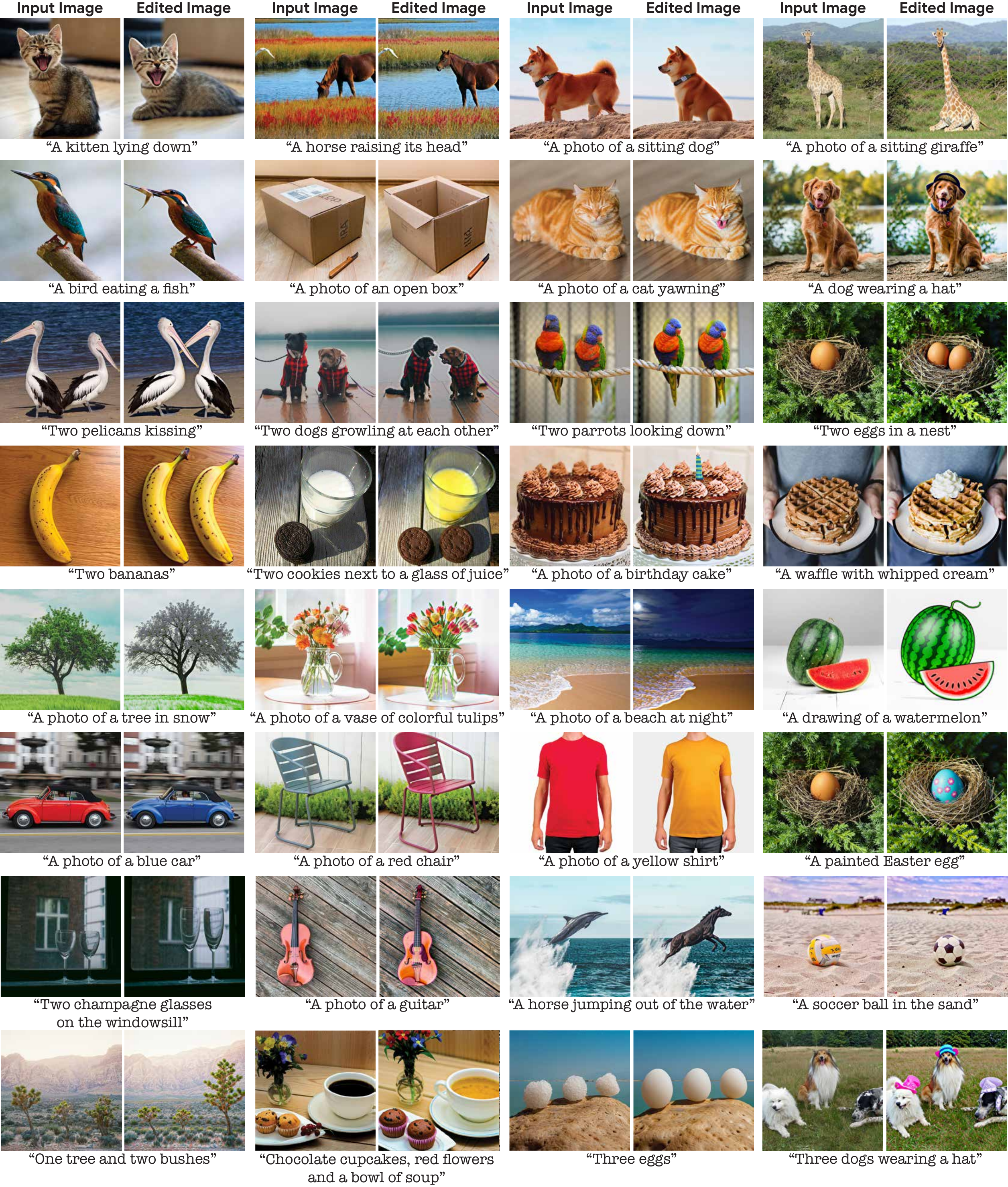}
    \caption{\textbf{Wide range of editing types.} { \it Additional $1024\times1024$-pixel pairs of original (left) and edited (right) images using our method (with target texts).
    Editing types include posture changes, composition changes, multiple object editing, object additions, object replacements, style changes, and color changes.}
    } %
    \label{fig:more_pics}
\end{figure*}
\vfill

\begin{figure*}
\vspace*{-0.1cm}
    \centering
    \includegraphics[page=1,width=0.95\textwidth]{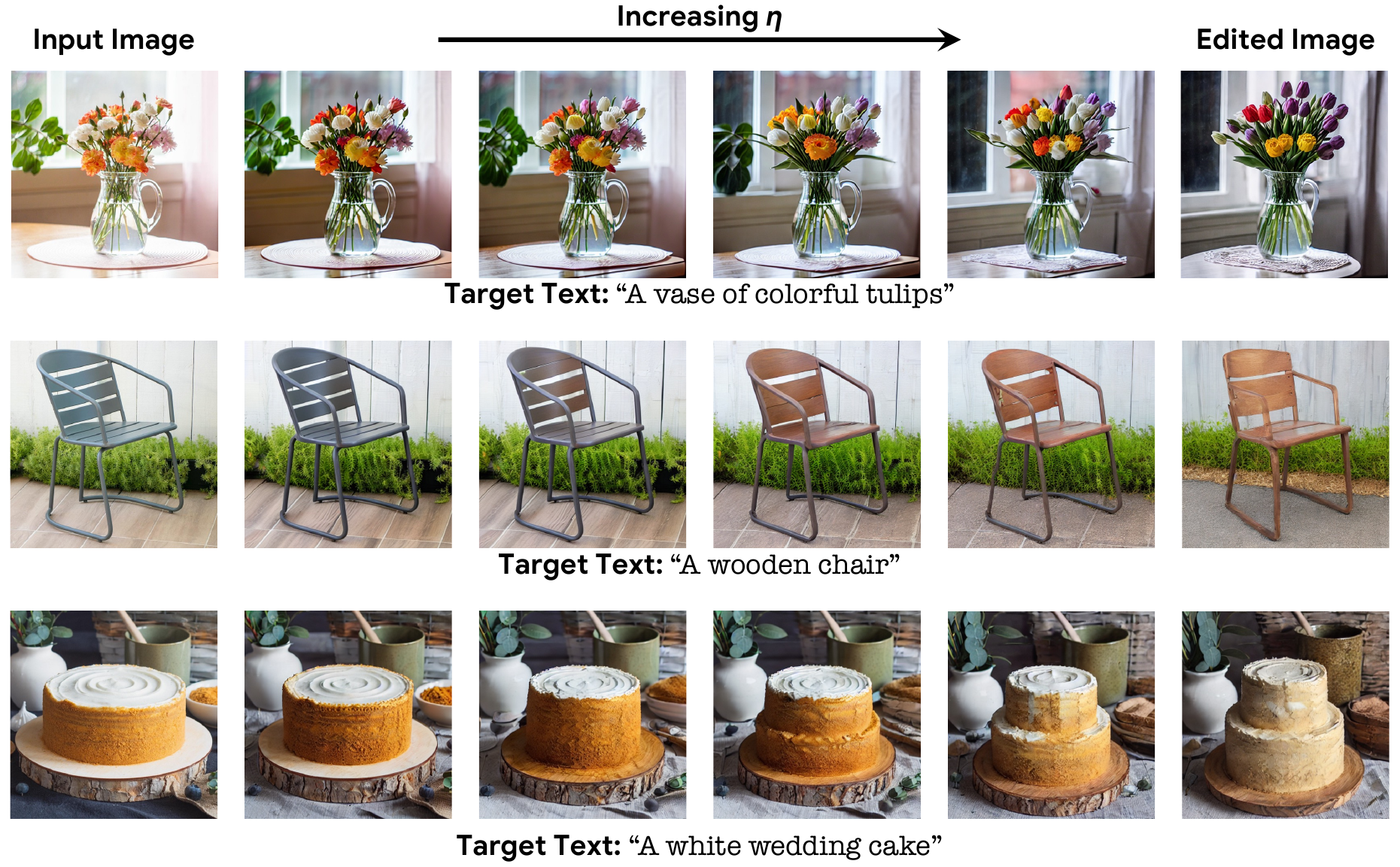}
    \vspace{-0.1cm}
    \caption{{\textbf{Smooth interpolation.} \textit{Additional results for smooth interpolation between the input image and the edited image using Imagic with Stable Diffusion (See animated GIFs in the supplementary material zip file).}}}
    \label{fig:smooth_Interpolation_more_results}
\end{figure*}

\begin{figure*}
\vspace*{-0.1cm}
    \centering
    \includegraphics[page=1,width=0.9\textwidth]{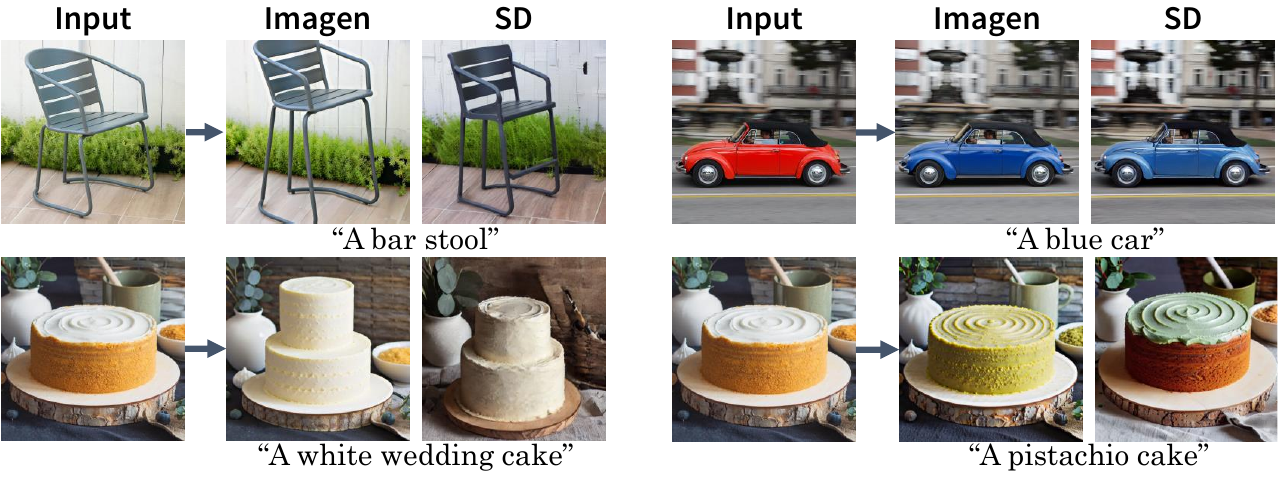}
    \vspace{-0.3cm}
    \caption{\textbf{Imagen vs Stable Diffusion.} \textit{Imagic's formulation is agnostic to the diffusion model choice. We show multiple examples of the same requested edit applied with either Imagen or Stable Diffusion.}}
    \label{fig:imagen_vs_sd}
\end{figure*}

\newpage
\section{Ablation Study}
In the paper, we performed ablation studies on model fine-tuning and interpolation intensity. Here we present
a discussion on the necessity of text embedding optimization, and
additional ablation studies on the number of text embedding optimization steps and our method's sensitivity to varying random seeds.

\paragraph{Text embedding optimization}
Our method consists of three main stages: text embedding optimization, model fine-tuning, and interpolation.
In the paper, we tested the value that the latter two stages add to our method.
For the final two stages to work well, the first one needs to provide two text embeddings to interpolate between: a ``target'' embedding and a ``source'' embedding.
Naturally, one might be inclined to ask the user for both a target text describing the desired edit, and a source text describing the input image, which could theoretically replace the text embedding optimization stage.
However, besides the additional required user input, this option may be rendered impractical, depending on the architecture of the text embedding model.
For instance, Imagen~\cite{imagen} uses the T5 language model~\cite{t5_transformer}. This model outputs a text embedding whose length depends on the number of tokens in the text, requiring the two embeddings to be of the same length to enable interpolation.
It is highly impractical to request the user to provide that, especially since sentences may have a different number of tokens even if they have the same number of words (depending on the tokenizer used).
Therefore, we opt not to test this option, and defer the pursuit of cleverer alternatives to future work.
Moreover, this dependence on the number of tokens prevents optimizing the model once per image, and then editing it for any text prompt.

\begin{figure}[t]
    \centering
    \includegraphics[width=\textwidth]{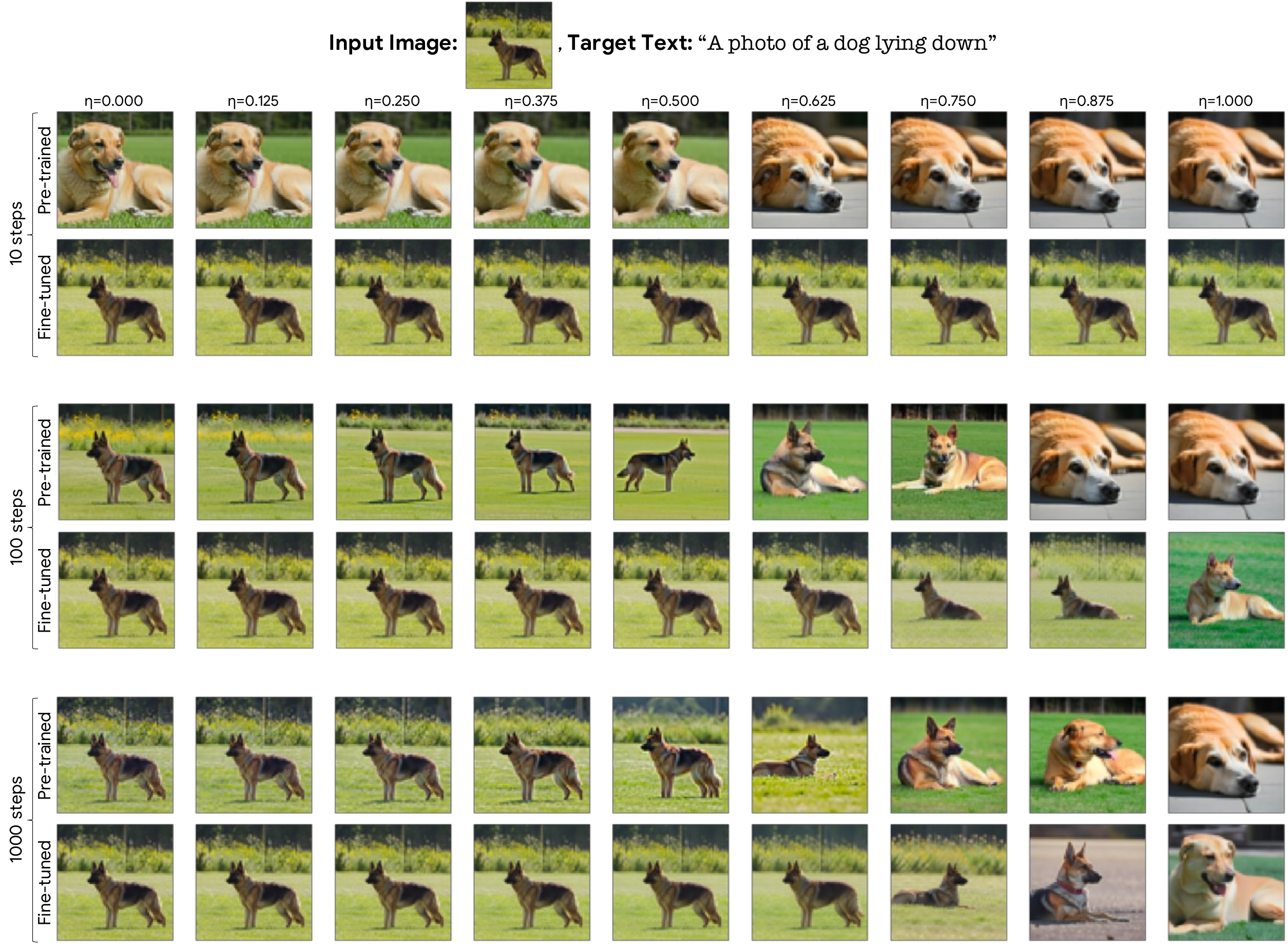}
    \caption{\textbf{Ablation for number of embedding optimization steps.} \textit{Editing results for varying $\eta$ and number of text embedding optimization steps, with and without fine-tuning (fixed seed).}} 
    \label{fig:appdx_steps_ft}
\end{figure}

\paragraph{Number of text embedding optimization steps}
We evaluate the effect of the number of text embedding optimization steps on our editing results, both with and without model fine-tuning.
We optimize the text embedding for $10$, $100$, and $1000$ steps, then fine-tune the $64\times64$ diffusion model for $1500$ steps separately on each optimized embedding.
We fix the same random seed and assess the editing results for $\eta$ ranging from $0$ to $1$.
From the visual results in \autoref{fig:appdx_steps_ft}, we observe that a $10$-step optimization remains significantly close to the initial target text embedding, thereby retaining the same semantics in the pre-trained model, and imposing the reconstruction of the input image on the entire interpolation range in the fine-tuned model.
Conversely, optimizing for $100$ steps leads to an embedding that captures the basic essence of the input image, allowing for meaningful interpolation. However, the embedding does not completely recover the image, and thus the interpolation fails to apply the requested edit in the pre-trained model. Fine-tuning the model leads to an improved image reconstruction at $\eta = 0$, and enables the intermediate $\eta$ values to match both the target text and the input image.
Optimizing for $1000$ steps enhances the pre-trained model performance slightly, but offers no discernible improvement after fine-tuning, sometimes even degrading it, in addition to incurring an added runtime cost.
Therefore, we opt to apply our method using $100$ text embedding optimization steps and $1500$ model fine-tuning steps for all examples shown in the paper.

\begin{figure}[t]
    \centering
    \includegraphics[width=\textwidth]{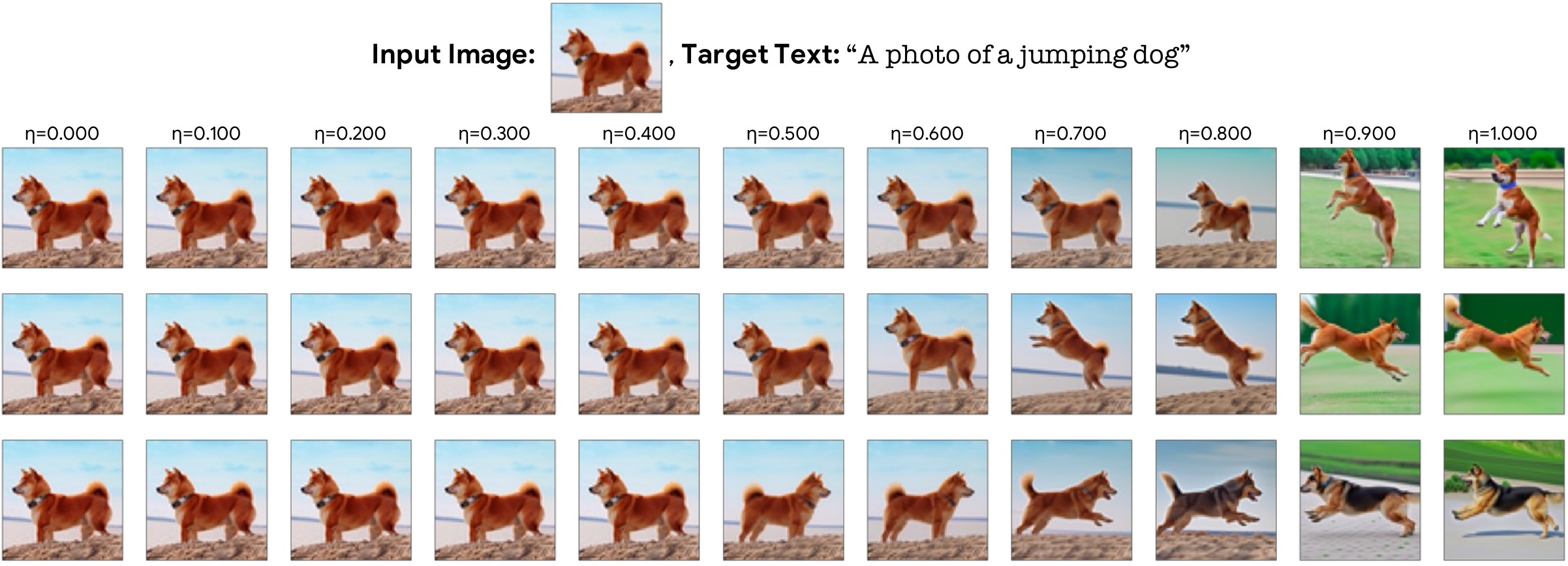}
    \caption{\textbf{Different seeds.} \textit{Varying $\eta$ values and different seeds produce different results for the same input.}}
    \label{fig:appdx_etas_seeds}
\end{figure}

\paragraph{Different seeds}
Since our method utilizes a probabilistic generative model, different random seeds incur different results for the same input, as demonstrated in \autoref{fig:seeds}.
In \autoref{fig:appdx_etas_seeds}, we assess the effect of varying $\eta$ values for different random seeds on the same input.
We notice that different seeds incur viable edited images at different $\eta$ thresholds, obtaining different results. For example, the first tested seed in \autoref{fig:appdx_etas_seeds} first shows an edit at $\eta=0.8$, whereas the second one does so at $\eta=0.7$.
As for the third one, the image undergoes a significant unwanted change (the dog looks to the right instead of left) at a lower $\eta$ than when the edit is applied (the dog jumps).
For some image-text inputs, we see behavior similar to the third seed in all of the $5$ random seeds that we test. We consider these as failure cases and show some of them in \autoref{fig:failures}.
Different target text prompts with similar meaning may circumvent these issues, since our optimization process is initialized with the target text embedding. We do not explore this option as it would compromise the intuitiveness of our method.

\section{User Study Details}
\begin{figure}
    \centering
    \includegraphics[width=0.95\textwidth]{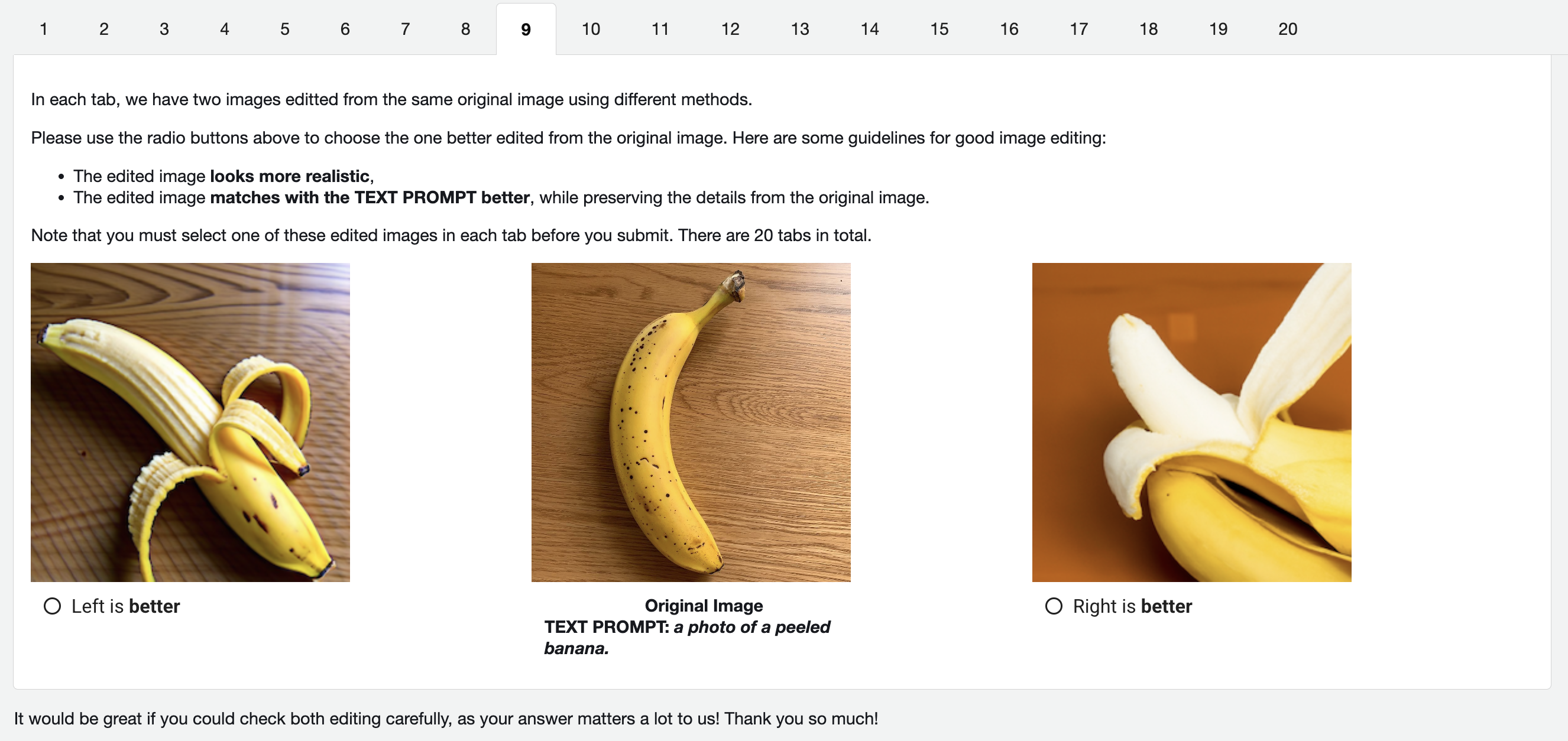}
    \caption{\textbf{User study screenshot.} \textit{An example screenshot of a question shown to participants in our human perceptual evaluation study.}}
    \label{fig:screenshot}
\end{figure}

We perform an extensive human perceptual evaluation study with \editbench\  ({Textual Editing Benchmark}), a novel benchmark containing $100$ image-text input pairs for the complex non-rigid image editing task.
The study was conducted using Amazon Mechanical Turk, to ensure unbiased evaluator opinions.
For each evaluator, we show a randomly chosen subset of $20$ images, including one image-text input pair that is shown twice. We discard all answers given by raters who answer the duplicate question differently, as they may not have paid close attention to the images.
Human evaluators were shown an input image and a target text,
and were asked to choose between two editing results:
A random result from one of SDEdit~\cite{sdedit}, DDIB~\cite{ddib}, or Text2LIVE~\cite{text2live}, and our result, randomly ordered (left and right).
Users were asked to choose between the left result and the right one, akin to the standard practice of Two-Alternative Forced Choice (2AFC)~\cite{kolkin2019style, park2020swapping, text2live}.
A sample screenshot of the screen shown to evaluators is provided in \autoref{fig:screenshot}.
We collected $3030$ answers for the comparison to SDEdit, $3131$ for DDIB, and $3052$ for Text2LIVE, totalling $9213$ user answers.

For fairness, we apply SDEdit, Text2LIVE, and \imagic\ using a single fixed random seed, while DDIB is deterministic and thus unaffected by randomness.
In \imagic, we choose the hyperparameter $\eta$ that applies the desired edit while preserving a maximal amount of details from the original image.
We choose SDEdit's intermediate diffusion timestep using the same goal. SDEdit was applied using the same Imagen model that we used, keeping its original hyperparameters.
We also apply DDIB using Imagen, with a deterministic DDIM sampler, an encoder classifier-free guidance weight of $1$, and a decoder classifier-free guidance weight ranging from $1$ to $5$ to control the editing intensity and choose the best result.
Text2LIVE is applied using its default provided hyperparameters.
Both DDIB and Text2LIVE had access to additional auxiliary texts describing the original image. 
The same hyperparameter settings were used in our qualitative comparisons as well.
It is worth noting that SDEdit, DDIB, and Text2LIVE were all designed without complex non-rigid edits that preserve the remainder of the image in mind. \imagic\  is the first method to successfully target and apply such edits.

Our results show a strong user preference towards \imagic, with all comparisons to baselines showing a preference rate of more than $70\%$.
We hope that \editbench\  enables comparisons in text-based real image editing in the future, and serves as a benchmark evaluation set for future work on complex non-rigid image editing.
To that end, we provide the full set of \editbench\  images and target texts along with results for all the tested methods at the following URL: \url{https://github.com/imagic-editing/imagic-editing.github.io/tree/main/tedbench/}.

\section*{Acknowledgements}
This work was done during an internship at Google Research.
We thank William Chan, Chitwan Saharia, and Mohammad Norouzi for providing us with their support and access to the Imagen source code and pre-trained models.
We also thank Michael Rubinstein and Nataniel Ruiz for insightful discussions during the development of this work.

\end{document}